\documentclass{egpubl}
\usepackage{VMV2023}
\usepackage{booktabs}
\usepackage{amsmath,amssymb}

\newcommand{\model}{MetaISP\xspace}

\newcommand\dE{\ensuremath{\Delta E^*_{ab} 76}\xspace}

\newcommand{\x}{\ensuremath{x}} %input image patch
\newcommand{\y}{\ensuremath{y^{*}_{d}}} %output image
\newcommand{\gtw}{\ensuremath{y^{w}_{d}}} %gt warped
\newcommand{\gt}{\ensuremath{y_{d}}} %gt warped
\newcommand{\dev}{\ensuremath{d}} %device
\newcommand{\emb}{\ensuremath{e_{d}}} %device
\newcommand{\wb}{\ensuremath{w_{d}}} %white balance

\ConferenceSubmission   % uncomment for Conference submission
\WsPaper           % uncomment for final version of EG Workshop contribution
\usepackage[T1]{fontenc}
\usepackage{dfadobe}  

\usepackage{cite}  % comment out for biblatex with backend=biber
\BibtexOrBiblatex
\electronicVersion
\PrintedOrElectronic
\ifpdf \usepackage[pdftex]{graphicx} \pdfcompresslevel=9
\else \usepackage[dvips]{graphicx} \fi

\usepackage{egweblnk} 
\usepackage{xspace}
\usepackage{hyperref}
\title[MetaISP -- Exploiting Global Scene Structure for Accurate
  Multi-Device Color Rendition]%
      {MetaISP -- Exploiting Global Scene Structure for Accurate
  Multi-Device Color Rendition}

\author[Matheus Souza \& Wolfgang Heidrich]
{\parbox{\textwidth}{\centering Matheus Souza\orcid{0000-0003-2751-0134}
        and Wolfgang Heidrich\orcid{0000-0002-4227-8508} 
        }
        \\
{\parbox{\textwidth}{\centering King Abdullah University of Science
and Technology, Saudi Arabia\\
       }
}
}

\begin{document}

\maketitle

\begin{abstract}
    Image signal processors (ISPs) are historically grown legacy software systems for reconstructing color images from noisy raw sensor measurements. Each smartphone manufacturer has developed its ISPs with its own characteristic heuristics for improving the color rendition, for example, skin tones and other visually essential colors. The recent interest in replacing the historically grown ISP systems with deep-learned pipelines to match DSLR's image quality improves structural features in the image. However, these works ignore the superior color processing based on semantic scene analysis that distinguishes mobile phone ISPs from DSLRs. Here, we present \model, a single model designed to learn how to translate between the color and local contrast characteristics of different devices. \model takes the RAW image from device A as input and translates it to RGB images that inherit the appearance characteristics of devices A, B, and C. We achieve this result by employing a lightweight deep learning technique that conditions its output appearance based on the device of interest. In this approach, we leverage novel attention mechanisms inspired by cross-covariance to learn global scene semantics. Additionally, we use the metadata that typically accompanies RAW images and estimate scene illuminants when they are unavailable. The source code and data are available at: \href{http://www.github.com/vccimaging/MetaISP}{github.com/vccimaging/MetaISP}.
%-------------------------------------------------------------------------
%  ACM CCS 1998
%  (see http://www.acm.org/about/class/1998)
% \begin{classification} % according to http:http://www.acm.org/about/class/1998
% \CCScat{Computer Graphics}{I.3.3}{Picture/Image Generation}{Line and curve generation}
% \end{classification}
%-------------------------------------------------------------------------
%  ACM CCS 2012
%   (see http://www.acm.org/about/class/class/2012)
%The tool at \url{http://dl.acm.org/ccs.cfm} can be used to generate
% CCS codes.
%Example:


\end{abstract}  
%-------------------------------------------------------------------------

\section{Introduction}
Mobile devices have emerged as the dominant platform for digital photography. The increasing demand for immediate photo sharing and consumption without manual post-processing has driven the need for advanced color processing capabilities in these devices. To meet this demand, mobile phones are now equipped with cutting-edge, high-level image processing techniques \cite{apple2021panoptic}, which greatly enhance color processing capabilities.

\begin{figure}[h!]
\centering
\includegraphics[width=0.9\linewidth]{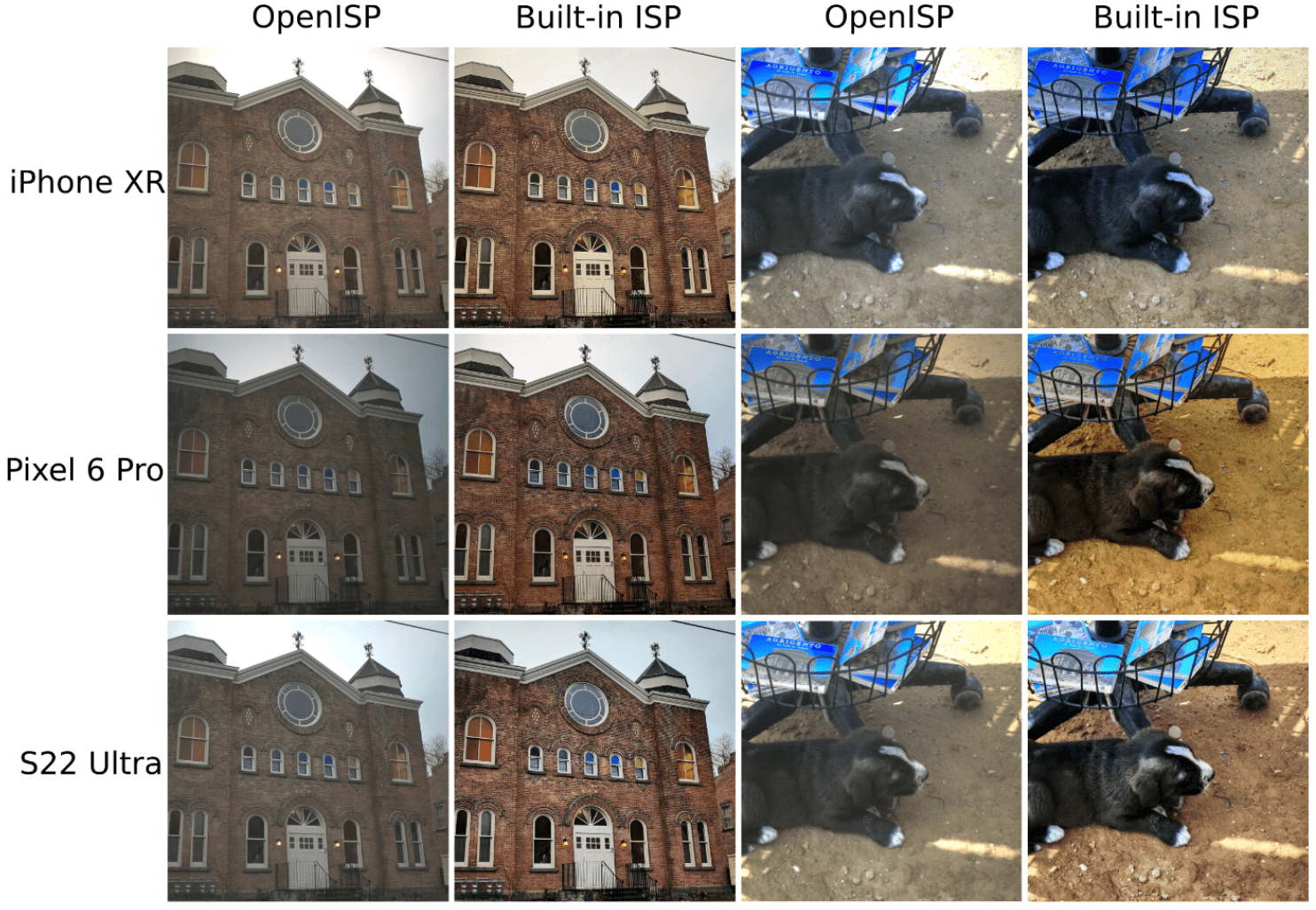}
\caption{Two scenes were captured using three different smartphone models and processed with different ISPs. When all RAW images were processed using the same OpenISP, the resulting appearance from all models was very similar. However, when the native ISPs for each device were used, strong appearance differences became visible.}

\label{fig:intro}
\end{figure}

In contrast, DSLR cameras still produce images with flat colors, placing the burden of color rendition on the end user. Over the past few decades, as camera modules have become smaller and image quality has suffered due to approaching physical limitations, mobile devices have increasingly relied on computational methods to restore high-quality images. The Image Signal Processor (ISP) in a mobile device plays a crucial role in this process. It translates the RAW sensor measurements into an sRGB image using various image processing stages, including interpolation, demosaicking, denoising, white balance, and color restoration. ISPs have accumulated extensive knowledge about color theory to generate visually pleasing images that not only match the scene accurately but also appeal to human perception, taking into account cultural differences.

Figure~\ref{fig:intro} highlights the importance of modern ISPs in overall image color reproduction. It presents images captured using three different smartphone cameras and processed using the (relatively simple) open-source OpenISP and the native ISPs of each device. Noticeably, there are strong color and shading differences in the images processed by the native ISPs, even though the images processed by OpenISP appear very similar (and much more "muted" in comparison). This demonstrates two points: 1) The final color reproduction on modern mobile phones is primarily determined by the ISP rather than the camera hardware; and 2) ISPs from different phones make distinct aesthetic choices regarding color reproduction, thereby establishing a brand-specific visual style.

In this work, we propose \model, an ISP that takes as input a RAW
image from one mobile phone and can closely reproduce the appearance
(color and local contrast characteristics) of any of a range of other
phones. This is achieved by training \model to learn to combine
metadata and global scene structure in a way that emulates the
existing black-box ISPs of these mobile phones. Since the different
device characteristics are encoded into a latent feature space, this
also makes it possible for \model to interpolate between different
device aesthetics.

\vspace{-0.3cm}
\section{Related Work}
\subsection{RAW Images and ISPs}

In recent years, there has been a growing interest in exploring the unprocessed RAW image space. RAW data, directly captured by the sensor, contains abundant information that has not been influenced or compressed by any Image Signal Processor (ISP). While sRGB images are widely available and commonly used, there have been several studies proposing the translation from sRGB to RAW, followed by performing enhancement tasks in the RAW domain, and finally converting the result back to sRGB \cite{zamir2020cycleisp,xing2021invertible,punnappurath2019learning,nguyen2016raw}. However, these approaches face certain limitations, primarily due to the inherent challenges in accurately reconstructing RAW images. ISPs are complex software components that are difficult to model and invert successfully.

Working directly in the RAW domain has led to significant advancements in various areas of image processing, including low-light image reconstruction, intrinsic image decomposition, image denoising, reflection removal, real scene super-resolution, illuminant estimation, and image enhancement \cite{chen2018learning,brooks2019unprocessing,abdelhamed2021leveraging,gharbi2016deep,lei2020polarized}. These methods often replicate parts of the ISP pipeline while leveraging deep learning techniques to match the output with a domain where the images are better conditioned. The idea of replicating an entire camera ISP was early explored by Karaimer and Brown \cite{karaimer_brown_ECCV_2016}, while Heide et al. \cite{heide2014flexisp} pioneered the replacement of numerous image processing blocks with a joint optimization approach. More recently, this approach has been extended to matching RAW images captured on mobile devices with high-quality DSLRs using deep learning. Ignatov et al. introduced the PyNet architecture \cite{ignatov2020replacing}, which utilized paired images from the Huawei P20 (RAW) and Canon 5D Mark IV DSLR (sRGB) to learn how to enhance the quality of mobile device images to match the DSLR level. Subsequently, several challenges \cite{contest2020,contest2021} have explored different architectures and strategies to achieve better translations between these domains.

Among the top-reported architectures, AWNet~\cite{awnet} and MW-ISPnet~\cite{contest2020} introduced Multilevel Wavelet CNNs and Residual Channel Attention Blocks. Subsequently, LiteISP~\cite{zhangliteisp} further improved upon these works by incorporating their successful techniques along with an alignment strategy based on optical flow and a color prediction network branch. Building upon these advancements, LiteISP proposed an architecture capable of faithfully reproducing the color of mobile ISPs. It leveraged raw metadata and generated high-resolution images as output. However, their approach was limited to learning one ISP at a time. In contrast, \model learns not only a multi-device ISP but also an illumination estimator, enabling both reproduction and interpolation between different device aesthetics.

\subsection{Image Translation}

The RAW-to-sRGB conversion can be interpreted as an image-to-image translation
problem. In recent years, much of the work in this area has focused on
solutions based on GANs~\cite{goodfellow2020generative}. The generative characteristic of those models
was explored to generate different solutions inside the same
domain, or conditioning the generated image to have different
characteristics based on a different domain input, which was called
multimodal image translation~\cite{zhu2017toward} and conditional image translation~\cite{pix2pix},
respectively. Later, these works inspired image reconstruction tasks like
colorization~\cite{larsson2016learning}, super-resolution~\cite{ledig2017photo}, inpaiting~\cite{pathak2016context}, etc. Recently,
works like All-in-One Weather Removal~\cite{li2020all} aim to remove
degradation caused by adverse weather conditions like rain, snow, or
fog with one single network. More recently,
TransWeather~\cite{valanarasu2022transweather} considered a similar
problem using a transformer-based network instead of
generative models. These works usually receive different corruptions as input and attempt to output a restored image.

Using a similar concept, we propose a multi-ISP single image reconstruction network to translate from one single RAW input to a desired sRGB image, conditioning the translations to a given label representing the target device, like early Generative Models \cite{mirza2014conditional}, but applying Convolution Neural Networks (CNNs), together with attention mechanisms as a global feature extractor.

\vspace{-0.3cm}
\section{Model}

\begin{figure*}[h]
\centering
\includegraphics[width=0.9\linewidth]{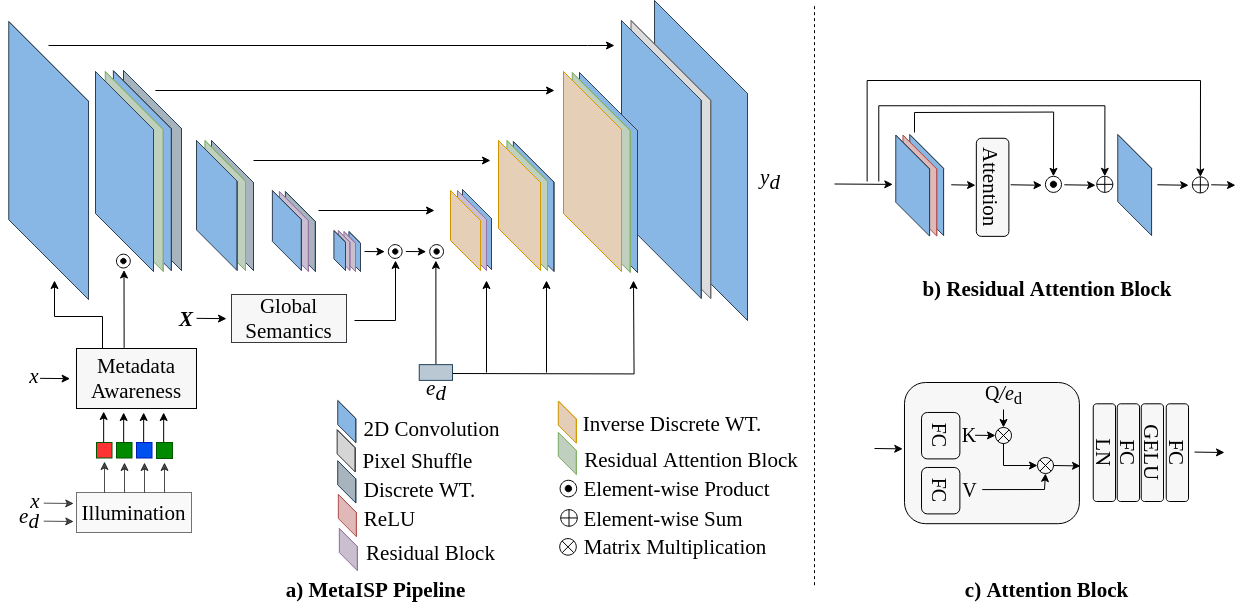}
\caption{a) is an overview of the \model pipeline, b) exemplifies the residual attention block, and c) describes the attention mechanism inside the residual blocks. In this architecture, the RAW patch image \x{} and \emb{} go through the illumination branch, outputting the scene white balance. Next, the metadata awareness block projects $WB_{d}$ to match the backbone network's first two levels and aggregate the ISO and exposure time. Later, a downsampled full-resolution version of the image goes through the transformer-based global semantics block to extract global features, matching them inside the bottleneck. Finally, the \emb{} conditions the decoding part with different device embeddings, serving as a scaling factor in the bottleneck and as the query vector inside the decoder attention mechanism. The number of convolutional layers, residual attention blocks, and residual blocks are the same as the Figure depicts. The residual block follows the b) diagram without the attention and the innermost skip connection.}
\label{fig:arch}
\end{figure*}

In this work, we proposed a RAW-to-RGB mapping that takes into account the possibility of multiple correct translations within different domains represented by three devices: Google Pixel 6 Pro ("Pixel"), Samsung S22 Ultra ("S22"), and iPhone XR ("iPhone"). The objective of our model is to reproduce the ISP of all these devices when provided with the same RAW image. To achieve this, our model is designed to handle small misalignments commonly found in datasets acquired using different phones. We accomplish faithful color reproduction by employing a CNN architecture that can produce diverse translations conditioned on a device embedding, incorporating a lightweight self-attention mechanism. Furthermore, we enhance our translations with a global semantics branch and a metadata awareness branch. The metadata awareness branch utilizes metadata information during training and can estimate it during inference, contributing to a more comprehensive translation process.

Given the raw RGBG image patch $\x \in\mathbb{R}^{\boldsymbol{H} \times \boldsymbol{W} \times 4}$, we aim to learn the target output $\y \in\mathbb{R}^{2\boldsymbol{H} \times 2\boldsymbol{W} \times 3}$. The device indicator $\dev$ represents the Pixel, S22, and iPhone with values 0, 1, and 2, respectively. During training, we select raw patches $\x$ from the iPhone and condition the network to replicate its original ISP or mimic the ISPs of the other two phones. For example, if $\dev = 1$, we match $\x$ with $y_1$.

The network should be able to encode general information from $\x$ while accurately decoding the specific details of the ground truth $\y$ based on the given device $\dev$. Furthermore, the network should be lightweight to ensure practical applicability. A detailed description of the specific model is provided below.

\subsection{Architecture}
\label{subsec:arch}
Among the architectures proposed in \cite{contest2020}, MW-ISPnet stands out by utilizing a Multilevel Wavelet UNet~\cite{unet} in combination with residual channel attention, resulting in significant improvements in the reconstruction process. Subsequently, LiteISP~\cite{zhangliteisp} demonstrated that it is possible to achieve high-quality results while reducing the number of attention residual blocks. \model builds upon this backbone architecture but incorporates modifications to tackle the challenging task of encapsulating different domains within a single model, while avoiding issues such as mode collapse or averaging of device appearances. Additionally, the network efficiently utilizes metadata and effectively extracts global semantics. For a comprehensive overview of the global architecture, refer to Figure~\ref{fig:arch}.

\subsubsection{Metadata awareness}
\label{sec:meta}
The raw file comprises unprocessed or minimally processed data and image metadata, which provides information about the characteristics of the image and the device settings chosen during the capture, considering factors such as lighting conditions and user preferences. While traditional ISPs make use of this data, most current Deep Learning models tend to overlook it. Within a DNG file, standardized metadata includes parameters like the color matrix, white balance weights, exposure settings, and ISO settings. 
\model considers the importance of white balance information when reproducing an ISP and incorporates two pipelines. In the first pipeline, where white balance information is available with the RAW image, the network assumes the white balance provided by the iPhone RAW image. Inside the network, the White Balance branch adapts this information to match the scene illuminants of the Pixel and S22 devices. The second pipeline, on the other hand, includes an Illumination Branch capable of estimating the white balance information given the RAW image and the device embedding.

In the first pipeline, the white balance metadata is represented as a scalar multiplier per color channel, with the green channel typically normalized to a value of 1. This information is expanded to $\wb \in \mathbb{R}^{4}$ in RGBG format. Convolutional layers then upsample this information to match the size of corresponding layers in the reconstruction backbone. This enables the white balance information to be injected into the reconstruction process at distinct locations through a channel-wise dot product. In the second pipeline, a lightweight model is employed to learn the white balance information when it is either unavailable or lost. In this case, the ground truth illuminant information from the collected dataset for the Pixel and S22 devices is applied. Therefore, the white balance can be approximated given the RAW image and the device. The network aims to capture global color information for each channel and follows the same procedure as the first pipeline.

Additionally, the model leverages the ISO and exposure time to learn an affine transformation described by the parameters $\alpha$ and $\beta$. This transformation globally alters the appearance of the image, ensuring even light distribution across the entire scene and effectively addressing vignetting issues(corners darker than the image center). The learning process involves projecting both the ISO and Exposure information independently into a higher dimension using linear layers followed by ReLU activation. These two vectors are then concatenated, and two additional linear layers are used to project them back to a lower dimension, matching the output dimension of the first white balancing branch procedure. The final two vectors ($\alpha$ and $\beta$) are used to scale and shift globally the image. Combining ISO and exposure time arises from the fact that both characteristics are related to global scene light condition.

\subsubsection{Conditioned reconstructions}
In addition to the RAW image $\x$, the model takes as input the device ID $\dev$. This ID is converted into an embedding $\emb \in \mathbb{R}^{128}$ using a lookup table. The embedding $\emb$ serves as a style vector that modulates the image, similar to the concept of style transfer in works like \cite{huang2017arbitrary,park2019arbitrary,liu2021adaattn} or image generation in \cite{karras2019style}. The embedding $\emb$ is then projected to a higher dimension using a convolution layer with a kernel size of 1, resulting in $\emb \in \mathbb{R}^{512}$. Finally, an elementwise product is performed to scale the latent space of the backbone network with the desired style. Among the various methods to modulate the bottleneck layer with style, the elementwise product is chosen as it is more suitable for performing inference with varying image resolutions compared to concatenation, which is commonly used in image translation tasks \cite{pix2pix,zhu2017toward}. The proposed method provides a constrained space for possible translations, similar to early conditional GANs \cite{mirza2014conditional}. This approach is well-suited for the goal of accurate color reproduction while also allowing for smooth interpolation between the styles of different devices by performing interpolation between the embedding vectors.

\subsubsection{Attention}
\label{sec:attention}
The backbone of \model is an autoencoder architecture that incorporates residual attention blocks after each convolutional layer. Inspired by works such as \cite{carion2020end,cui2022light,liu2021adaattn,valanarasu2022transweather}, we designed an attention mechanism that utilizes a global query vector during encoding and a style-based query vector during decoding.
In the encoder, each attention block is initialized with an embedding query vector $Q \in \mathbb{R}^{dim}$, where $dim$ represents the dimension at the current level. This vector is expanded to match the batch size and multiplied by the keys $K \in \mathbb{R}^{HW \times dim}$. The softmax function is then applied to calculate the attention weights, which are multiplied with the values $V \in \mathbb{R}^{HW \times dim}$.

On the other hand, the decoder receives the style embedding vector, which is projected using a 1x1 convolution to match the dimension of the current level. This projected embedding serves as the query $Q$ in the attention mechanism and undergoes the same attention computation steps as in the encoding phase. Here, $h$ represents the number of attention heads. The keys and values are calculated using fully connected (FC) layers applied to the image's encoded features, which are normalized by LayerNorm.
After the attention calculation, the attention map is multiplied with the image in a residual fashion and passed through FC layers, LayerNorm, and GELU activation, while also applying a residual connection. This approach suits our purpose because all devices share the same RAW input. We aim to learn a model that effectively encodes the RAW image and combines the bottleneck with the style features during decoding, thereby enforcing the individual style of each device within the attention mechanism.

The attention layer is integrated into the residual attention block, as depicted in Figure~\ref{fig:arch}. For each attention computation, we incorporate positional encoding, where each patch has different coordinates representing its position on the full-resolution image. Notably, transitioning from channel attention to the suggested approach necessitates just a single attention block, and this change is applied in only a select few layers.

\subsubsection{Global semantics}
Relying solely on local patches is insufficient as they lack information about the relationships between different objects in the scene. Patches only provide texture or cropped structure observations. To address this limitation, \model incorporates a cross-covariance attention mechanism, utilizing a downsampled version of the entire image $\x$ pre-processed through bilinear interpolation.
The cross-covariance transformer, introduced by \cite{ali2021xcit}, offers a lightweight attention mechanism by computing self-attention with respect to the feature dimension. We leverage this transformer to aggregate information from the entire image, exploiting the known large receptive field property. The final representation of the transformer is obtained using the CLS layer \cite{cls}, which is commonly employed in classification tasks to aggregate information on the whole image. The features are then projected using a convolutional layer with a kernel size of 1 to match the dimension of the bottleneck layer. Finally, the projected features are merged with the bottleneck using element-wise multiplication. This approach aligns with current commercial ISPs, which often perform rough classifications, and the model implicitly learns which features are more important in the scene.

\medskip\noindent\textbf{Normalization} Transformers typically require large amounts of data for effective training. Although the cross-covariance approach helps reduce computational complexity, it still tends to require significant data resources compared to pure CNNs. In this work, we proposed the use of monitor data augmentation, as described in Section~\ref{sec:data}. Initially, \model is pre-trained using monitor-captured data to increase its semantic diversity. During the subsequent fine-tuning procedure, we retain the statistics learned from the monitor data in the global semantics stage. This means that during fine-tuning, the batch normalization does not compute new mean and variance based on the current mini-batch but instead uses the running mean and variance learned during the pre-training with monitor data. Section~\ref{sec:res} provides a quantitative evaluation of this strategy.

\medskip\noindent\textbf{Light Transformer} Our global semantics branch divides the image in $16x16$ patches, with $4$ blocks with attention dimensionality of $128$ and $4$ heads, making it smaller than the tiny version proposed in \cite{ali2021xcit}. With its additional 1.39M parameters, the model is still lightweight compared with other methods.

\vspace{-0.3cm}
\section{Dataset and Training Details}
\label{sec:data}

In the existing literature, there is a lack of large datasets that provide aligned images from different mobile devices. Although PyNet~\cite{ignatov2020replacing} offered aligned Huawei RAW patches with DSLR images, it did not provide the opportunity to explore strategies like the ones proposed in this work (refer to Section~\ref{subsec:arch}). Furthermore, DSLR images, unlike current mobile devices, typically require post-processing to achieve good color rendition.

To overcome these limitations, we conducted a data collection process to obtain real-world images from each device. Our dataset consists of 163 diverse scenes captured in indoor, outdoor, night, and daily environments. In addition to the iPhone XR, Google Pixel 6 Pro, and Samsung S22 Ultra, we also collected images from the Xiaomi C40 to evaluate the model in a zero-shot fashion. This resulted in a total of 652 images, which were further divided into patches as described below. During the data collection process, all phones were set to automatic mode for white balance, focus, ISO, and exposure time. The captures were performed using a tripod, with the phones arranged in a grid formation, and a remote controller was used to trigger simultaneous shots.

To enrich our dataset, we also captured images in front of a calibrated HDR monitor. This data was exclusively used for pre-training our model, allowing it to learn semantics from a broader range of scenes and multiple devices. The images were captured in a dark room using the Eizo CG3145 4K HDR monitor, which was carefully color-calibrated to ensure accurate representation and the use of an HDR monitor means that we can produce images that exceed the contrast of the camera so that we obtain realistic image saturation and glare effects. The reference images in the monitor dataset were sourced from Flickr2K~\cite{Lim_2017_CVPR_Workshops} and cover a wide range of landscapes, vegetation, objects, indoor and outdoor scenarios, etc. The monitor dataset consists of a total of 5487 RAW/sRGB image pairs with a resolution of $4032\times3024$. The impact of the pre-training strategy is quantitatively demonstrated in Section~\ref{sec:res}. We will make the entire dataset, including the pre-training and real-world images, publicly available.

\begin{table*}[h]
\centering
\caption{Quantitative comparison of MetaISP and the illuminants variation against state-of-the-art methods evaluated using patches.}
\begin{tabular}{@{}lrrrrrrrrr@{}}
\toprule
Model     & \multicolumn{3}{c}{iPhone XR} & \multicolumn{3}{c}{Pixel 6 Pro} & \multicolumn{3}{c}{Samsung S22} \\ \midrule
          & PSNR $\uparrow$    & \dE $\downarrow$    & SSIM $\uparrow$    & PSNR $\uparrow$     & \dE $\downarrow$   & SSIM $\uparrow$     & PSNR $\uparrow$     & \dE $\downarrow$     & SSIM $\uparrow$    \\ \midrule
MW-ISPnet~\cite{contest2020} &   26.32 &    5.84    &  0.892      &  22.39   &   9.18    &  0.659   &  20.92   & 9.88 & 0.639        \\
LiteISP~\cite{zhangliteisp}   &   26.87 &    5.54    &  0.903 &   22.97  &    9.03    &   0.759 &   21.90  &  9.46      &   0.726  \\ 
SwinIR~\cite{liang2021swinir}    &   25.76      &    5.92    &  0.889      &   21.75      & 9.22   &   0.637 &  20.49       &     9.98& 0.627        \\ \midrule
MetaISP & \textbf{30.14}   &    \textbf{3.76}    &  \textbf{0.915} &   \textbf{25.49}      & \textbf{6.65    }  &  \textbf{0.784}        &   \textbf{24.26}  & \textbf{7.33}   &\textbf{ 0.753 }        \\ 
I-MetaISP & 29.12 & 5.07 & 0.897 &    24.89 &  7.28  & 0.780  & 23.66 & 7.77  & 0.749       \\  \midrule
\end{tabular}

\label{tab:results}
\end{table*}

\vspace{-0.3cm}
\section{Results}
\label{sec:res}

\begin{figure*}[h!]
\centering
\includegraphics[width=0.8\linewidth]{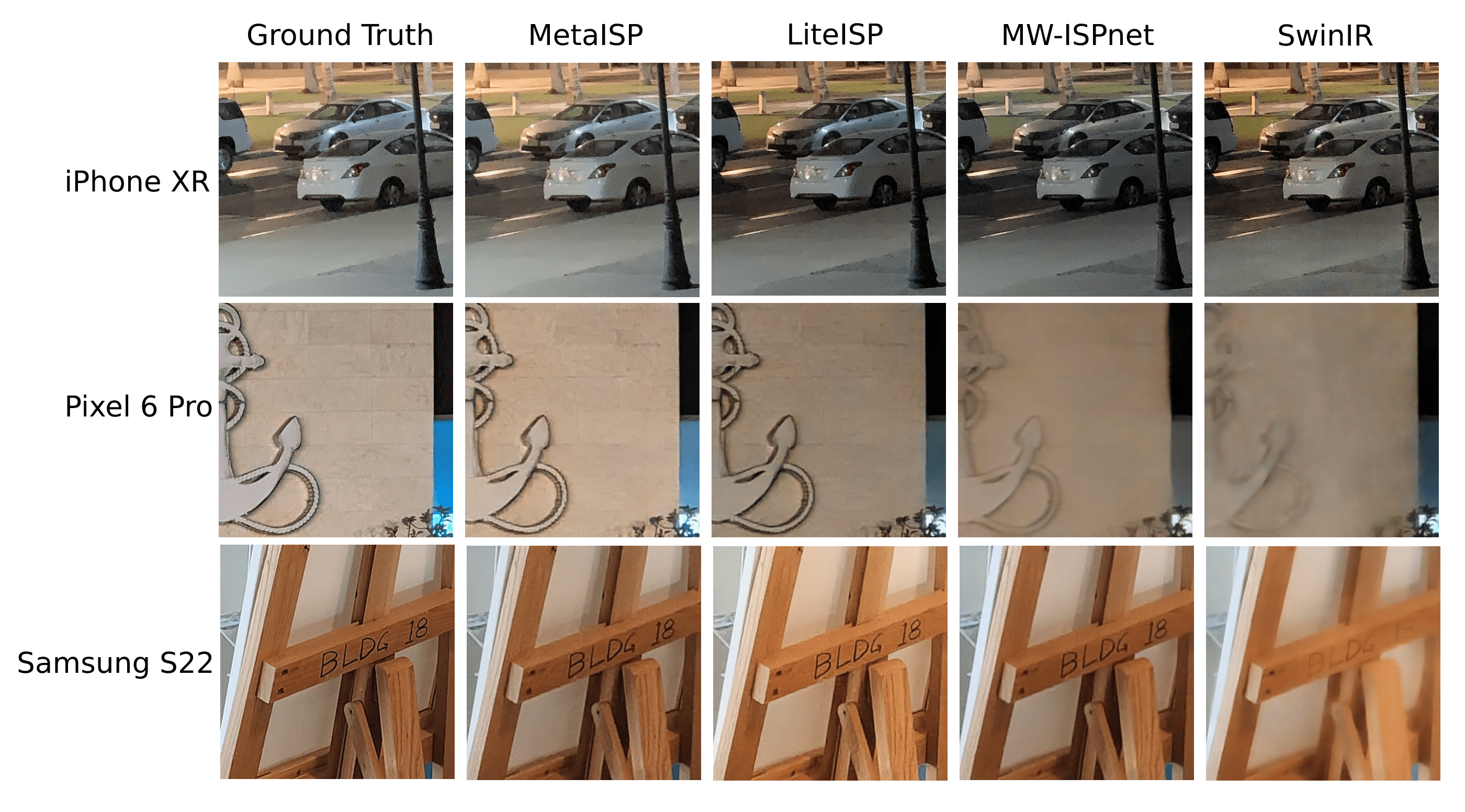}
\caption{Patch-wise comparison between the different methods and the ground truth. The images mainly represent challenging scenarios, such as indoor and night scenes. Our model can accurately reproduce the color appearance of each target device. The alignment strategy described in the supplementary effectively avoids blurry effects that may be observed in SwinIR, resulting in sharper and more visually pleasing results. On the other hand, LiteISP does not suffer from blurry effects but struggles reproducing the color perception.}
\label{fig:overview}
\end{figure*}

To assess the performance of our model, we conduct a quantitative comparison with state-of-the-art methods using our collected datasets, performing the pre-training with monitor-captured data and training with the proposed real-world dataset. Furthermore, we perform a qualitative comparison of all methods against ground-truth images. Additionally, we demonstrate the capability of our model to interpolate between different device styles, as well as its potential to process images from phone models that were not included in the training data, albeit with reduced color accuracy.

Our model is comapred against the state-of-the-art methods LiteISP~\cite{zhangliteisp}, MW-ISPNet~\cite{contest2020}, and SwinIR~\cite{liang2021swinir}. These works primarily focus on image enhancement by translating RAW to sRGB. SwinIR, originally designed for image super-resolution, was adapted to handle four-channel raw images. As no previous method has addressed multi-ISP before, we trained each technique separately to match the quality of the three devices in our dataset.

The raw input image for all methods is the iPhone image multiplied by the white balance. However, in our case, we utilized the white balance adaptation strategy described in Section~\ref{sec:meta}. Our experiments confirm this strategy's contribution to improving our model's performance (see Table~\ref{tab:ablation}).

For quantitative evaluation, we adopted commonly used metrics, including PSNR (Peak Signal-to-Noise Ratio), SSIM (Structural Similarity Index), and \dE (perceptual color accuracy error). These metrics were measured using ground-truth images that were warped to align with each network's output using PWC-Net~\cite{sun2018pwc} for all methods. This alignment step was performed to mitigate the contribution of misalignment in the metrics, allowing us to focus on evaluating color accuracy. Please check the supplementary material for more details regarding the alignment.

Table~\ref{tab:results} demonstrates that \model outperforms state-of-the-art methods in all the evaluated metrics and across different devices. In an all-in-one fashion, our method can better reconstruct the target devices without having to train individually for each task (refer to supplementary for ablation about this case). Table~\ref{tab:results} also highlights that when the illuminants are learned together with the model (\textit{I-MetaISP}), the performance is slightly worse. This is attributed to the challenge of accurately estimating illuminants, as minor deviations can result in different color temperatures, thereby changing the image's overall appearance. However, we still consider this approach to help compensate for cases where illuminant information is not available. In some cases, it performs better when targeting Pixel and Samsung color appearances. This can be explained by the fact that, in this approach, we train the model to match their original illuminants, which could be more accurate than iPhone (original RAW source) values adapted by the white balance branch in the network. 

\begin{table*}[]
\centering
\caption{Ablation over the different contributions. Each one of them has a letter from A to E.}
\begin{tabular}{@{}lrrrrrrrrr@{}}
\toprule
Configuration    & \multicolumn{3}{c}{iPhone XR} & \multicolumn{3}{c}{Pixel 6 Pro} & \multicolumn{3}{c}{Samsung S22} \\ \midrule
          & PSNR $\uparrow$    & \dE $\downarrow$    & SSIM $\uparrow$    & PSNR $\uparrow$     & \dE $\downarrow$   & SSIM $\uparrow$     & PSNR $\uparrow$     & \dE $\downarrow$     & SSIM $\uparrow$    \\ \midrule
A Baseline  &   25.29 &   6.73   &  0.870 &   22.44  &  9.84  &  0.757 &   21.23  &  10.36 &   0.723  \\ 
B + Adapt. Illuminants   &   25.30 & 6.30  &0.858    &   22.48  &  9.49  & 0.744  &   21.33  & 9.79 & 0.717   \\ 
C + Global Semantics & 27.23   &   4.88    &  0.887 &  24.44   & 7.33       &  0.767       &   23.59  & 7.65   & 0.735        \\ 
D + Attention & 29.44 & 4.02 & 0.904 &   24.82   &  7.07  & 0.775   &  24.21 &\textbf{7.10}  & 0.743        \\
E + Iso and Exp & \textbf{30.14}   &    \textbf{3.76}    &  \textbf{0.915} &   \textbf{25.49}      & \textbf{6.65}      &  \textbf{0.784}        &   \textbf{24.26}  & 7.33   & \textbf{0.753}         \\  \midrule
\end{tabular}

\label{tab:ablation}
\end{table*}

The metrics also evidence the challenging task of ISP reproduction. State-of-the-art image reconstruction networks cannot perform as well as models designed to treat this problem, LiteISP~\cite{zhangliteisp} and MW-ISPNet~\cite{contest2020} explore domain-specific knowledge about the ISPs to provide mechanisms that improve the raw-to-rgb reconstruction, which lacks in SwinIR~\cite{liang2021swinir}. \model combines novel deep-learning architectures and insights from legacy ISPs.

Qualitatively, Figure~\ref{fig:overview} visually demonstrates the closeness of our \model results to the ground truth. With a scalar combined with the RAW file as input, \model can generate a diverse range of color perceptions. In contrast, the comparison methods require training from scratch to match the appearance of a specific device, and still, they fall short. SwinIR~\cite{liang2021swinir} and MW-ISPNet~\cite{contest2020} lack a mechanism to compensate for misalignments in the dataset, resulting in blurry outputs. LiteISP~\cite{zhangliteisp} has such feature, but it struggles with accurate color reproduction, which is crucial for reproducing the color rendition of mobile devices. Please check the supplementary material for more visualizations.

\begin{figure}[h]
\centering
\includegraphics[width=1.0\linewidth]{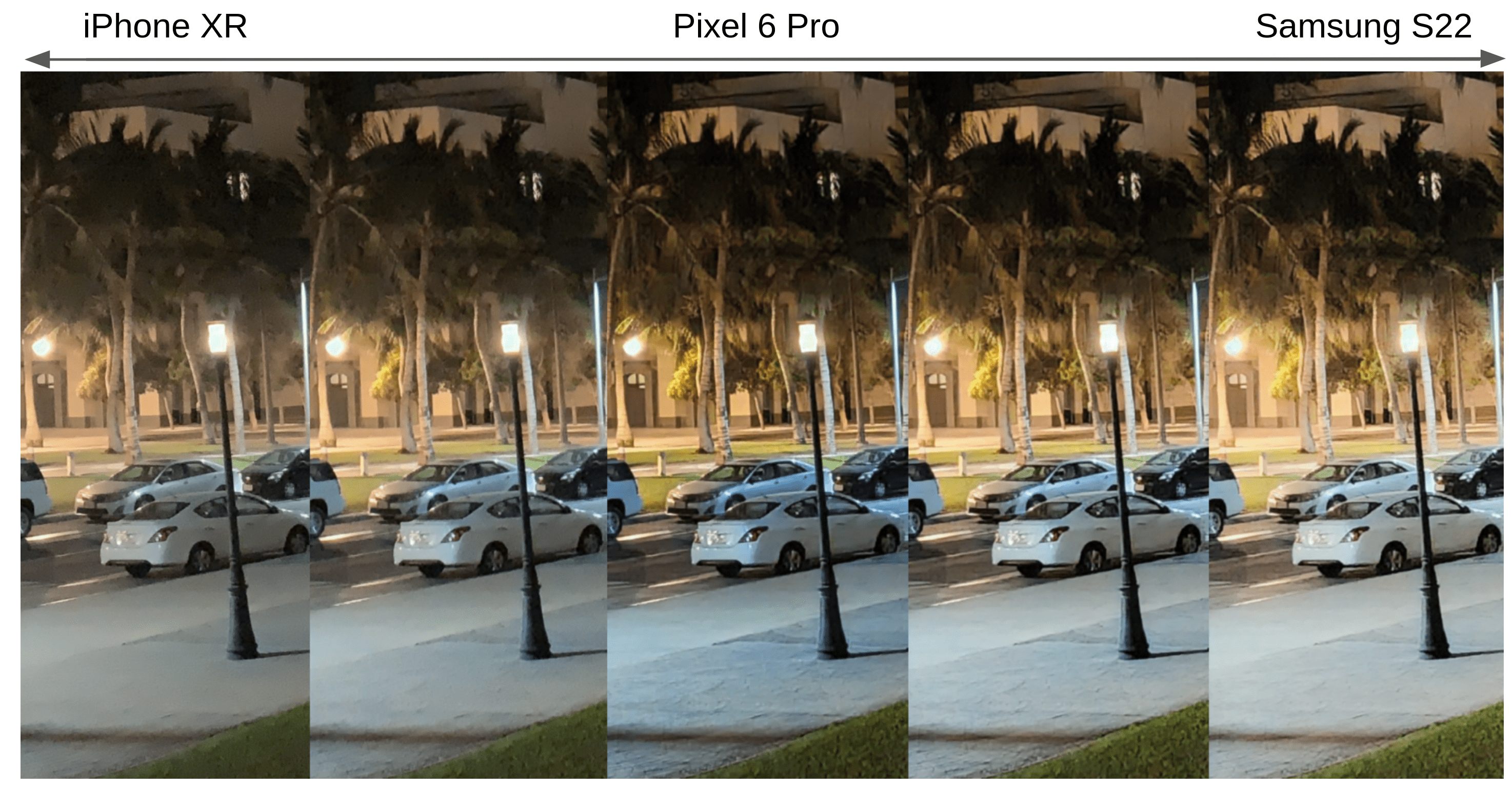}
\caption{Here, we exemplify the possibility of interpolating between the embeddings of different devices generating intermediate images. We can notice how different ISPs understand the scene and put more emphasis on different features. We have iPhone XR, Pixel 6 Pro, and S22 Ultra from left to right.}
\label{fig:lw1}
\end{figure}

\medskip\noindent\textbf{Interpolation} 
Another feature of our work is the interpolation between different device embeddings, producing appealing intermediate results. Figure~\ref{fig:lw1} shows that possibility with one intermediate result between each device. This number can be increased, creating very smooth transitions. 

\begin{figure}[h]
\centering
\includegraphics[width=1.0\linewidth]{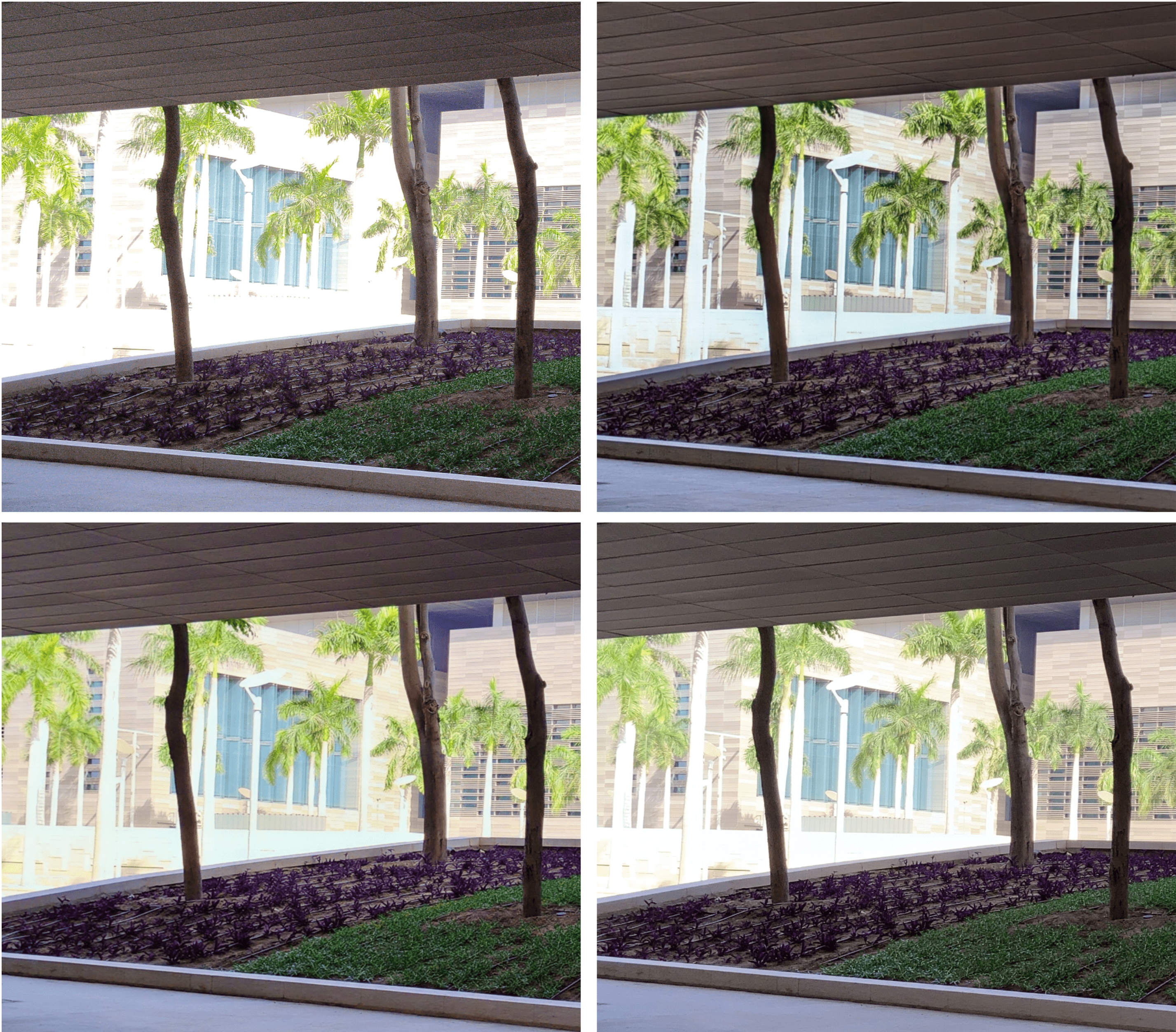}
\caption{Zero-shot processing of images from new devices.
  \model generating different interpretations from a Xiomi C40
  input without additional training. Top-left Xiomi C40 (input,
  top-right simulated Pixel 6 Pro, bottom-left simulated S22 Ultra, and
  bottom-right simulated iPhone XR. Please zoom in for more details.}
\label{fig:xiomi}
\end{figure}

\medskip\noindent\textbf{Zero-shot qualitative evaluation} Figure~\ref{fig:xiomi} shows how the reconstruction looks when another smartphone is considered without training. On the top-left corner, we have the Xiaomi C40 sRGB, and the other three images are the reconstructions when its RAW image passes through the model. We can notice an overall improvement, with crisp colors, better contrast, and more fine details compared to the native ISP of this inexpensive phone. Please see the supplement for more results.
\vspace{-0.3cm}
\subsection{Ablation Studies}

In our work, we have provided a detailed quantitative evaluation for each contribution, as outlined in Table~\ref{tab:ablation}.

A represents the baseline, a network similar to LiteISP but without channel attention. However, this baseline produces slightly inferior results. Next, we added B, the white balance first pipeline as described in Section~\ref{sec:meta}. This pipeline learns how to retain the white balance information, injecting it into deeper layers while also adapting the values for different phones' color perception instead of only just pre-multiplying it by the white balance scalars. C represents the global semantics branch, which provides knowledge about the global relationship between image patches. This semantic information is injected into the bottleneck of the main component, and its success is attributed to the normalization process that heavily depends on the pre-training strategy. The pre-training procedure learns global semantics from a diverse set of scenarios provided by the monitor data, and the normalization adopted retains the statistics of this first process while training with real-world images.  
D is the addition of the attention mechanism as described in Section~\ref{sec:attention}. 
This strategy enables the learning of global features within the patch and serves as a conditioning mechanism using the query vector from the embedding form of the device. Finally, E involves using the exposure time and ISO value to help control global light scene characteristics as described in Section~\ref{sec:meta}.

\section{Conclusion}
In this work, we have successfully built a single model to reproduce
the complex color and appearance decisions performed by the ISPs of three commercial devices. Even though each smartphone has a different sensor, lenses, etc., the ISPs are responsible for the brand-specific visual style, and \model learns jointly how to perform this task without collapsing or averaging to one appearance. With fewer parameters (see supplementary) and performing three tasks at once, the proposed method outperforms the current state-of-the-art in all metrics (PSNR, SSIM, and \dE), enhancing the original image or creating new appearance possibilities, interpolating between different phone style spaces. This work also exploits the white balance information that usually comes with RAW files, and it's also robust to cases where this information is not available, learning how to estimate it using a light model that receives the RAW and device embedding as inputs. We also propose a novel idea to extract global semantics extra-patches and attention to account for global communication intra-patches.

Finally, in the future, we want to explore different devices, understand their characteristics, learn how to translate from other devices without any further training requirements, expand the idea of zero-shot presented here, and add more control over the styles.

%-------------------------------------------------------------------------

\bibliographystyle{eg-alpha-doi}

\bibliography{MetaISP/references}
 
\setcounter{section}{0}
\setcounter{equation}{0}
\setcounter{figure}{0}
\setcounter{table}{0}
\setcounter{page}{1}
\onecolumn
\begin{center}
    {\LARGE\textbf{Supplementary Material}}\\[0.5cm] % Supplement title
    \medskip
    {\large\textbf{MetaISP -- Exploiting Global Scene Structure for Accurate
  	Multi-Device Color Rendition}}\\[1cm] % Optional subtitle
  	\medskip
\end{center}

\section{Training Details}
The images are divided into 1463, 186, and 180, respectively, for training, validation, and testing for each device during the pre-training. During the effective training, the division is 137 for training and 26 for testing.

The same pictures are available for all devices and were pre-aligned, as described below. To avoid memory issues, during training, the data is divided into patches. First, the images are cropped to $2688\times2688$ resolution. Each image is divided into 36 patches with a resolution of $448\times448$. Resulting in a final dataset size of 65844 for pre-training and 5868 for training.

\model considered as input the iPhone RAW image and, as possible, output the Google Pixel 6 Pro, Samsung S22 Ultra, and the iPhone XR
sRGB images. Hence, the RAW from Pixel and S22 is not used, only their metadata. Therefore, when learning the white balance, we can match the specific values of each device. When the white balance is not learned, the metadata is taken from the source device (iPhone), since the metadata of the target model is not available during inference. This was
empirically shown to produce better results than not using metadata.

Our model is training for 100 epochs, with a learning rate (LR) of $10^{-4}$, ADAM optimizer, and a scheduler decreasing the LR linearly to 0 after half the epochs. During training, we randomly selected the device ID and its ground truth. Hence, different mobiles are considered in one single minibatch. Horizontal and vertical flips are used as augmentations.

\section{Alignment and Pre-processing}
\label{sec:allign}

After making efforts to capture the aligned data, misalignments are still present due to the grid shape in which the phones were arranged during the capture process. To address this issue, we initially perform rough alignment using the RANSAC~\cite{vedaldiransac} algorithm, aligning the RGB images from the Pixel and S22 phones with the RGB image from the iPhone. Consequentially, we align them with the corresponding raw images, as the iPhone RGB and its raw images are already perfectly aligned. However, this may not be sufficient, as it can result in blurry outputs and color inconsistencies during training. To overcome this challenge, we employ optical flow during training to align all the ground truth images with the iPhone RGB image. This technique was initially introduced by Zhang et al.~\cite{zhangliteisp}, who proposed a module for matching raw image with rgb images color, then warp the original ground truth images to the raw-colored. In our case, we simplify this process by utilizing the iPhone RGB image as the reference for aligning the Pixel and S22 images, eliminating the need for an additional module (see Equation~\ref{eq:gt}. Furthermore, we implement a pixel occlusion technique, as suggested by \cite{meister2018unflow}, where pixels exhibiting a mismatch between forward and backward optical flows are occluded.

\begin{equation}\label{eq:gt}
\gtw = \mathcal{W}(y_{2}, \gt),
\end{equation}

Where $\dev \in \left \{ 0,1 \right \}$ and $\mathcal{W}$ is the warping operation, using the flow computed by the pre-trained PWC-Net~\cite{sun2018pwc}.

On the raw side, we need to pre-process to better position our input and have a "neutral" raw space. Hence, we normalize the image using the white and black level information in the metadata, which also provides white balance, ISO, and exposure time.
\section{Loss Function}

As the ground-truth images are aligned using the optical flow from a pre-trained PWC-Net \cite{sun2018pwc}, we need to mask out pixels that were removed from the scene to avoid color inconsistencies. Therefore our losses follow Equation~\ref{eq:loss}.

\begin{equation}
\label{eq:loss}
\begin{matrix}
\mathcal{L}_{L1}(\y,\gtw) &=& \left \| m\cdot (\y-\gtw) \right \|_1  \\
\mathcal{L}_{VGG}(\y,\gtw) &=& \left \| m\cdot (\y-\gtw) \right \|_1  \\
\mathcal{L}_{SSIM}(\y,\gtw) &=& \left \| m\cdot (\y-\gtw) \right \|_1
\end{matrix}
\end{equation}
where $m$ is the mask calculated based on the correct optical positions. We occluded pixels with flow $ < 0.999$. We also considered occluding pixels with a mismatch between forward and backward flows as proposed by \cite{meister2018unflow}. 
In Equation~\ref{eq:loss}, L1 is the mean absolute error, VGG is the perceptual loss calculated with features extracted from a VGG-19 \cite{vgg} pre-trained model, and SSIM \cite{ssim} stands for the structural similarity index. Perceptual and structural loss is even more critical in reconstruction tasks when any source of misalignment can be found. The total loss is described by Equation~\ref{eq:total}.

\begin{equation}
\label{eq:total}
\begin{matrix}
\mathcal{L}_{Total}(\y,\gtw) =& \lambda_{L1}\mathcal{L}_{L1}(\y,\gtw)\\
& + \lambda_{VGG}\mathcal{L}_{VGG}(\y,\gtw)\\
& + \lambda_{SSIM}\mathcal{L}_{SSIM}(\y,\gtw) , 
\end{matrix}
\end{equation}
where $\lambda_{L1}$,$\lambda_{VGG}$ and $\lambda_{SSIM}$ are respectively 1,1, and 0.1.

Our proposed strategy also learns the white balance with the image translation task. Therefore, another loss element should be added to account for this. We observed that training the illumination branch sharing the same optimizer as the reconstruction network produces better results. Equation~\ref{eq:illu} shows the final loss structure.

\begin{equation}
\label{eq:illu}
\begin{matrix}
\mathcal{L}_{TotalWB}(\y,\gtw,WB_{dg}) = \mathcal{L}_{Total}(\y,\gtw) \\
 + \lambda_{illu}\left \| (WB_{d}-WB_{dg}) \right \|_1  , 
\end{matrix}
\end{equation}
where $\lambda_{illu}$ is equals to $0.1$.

\section{Additional Ablations}

\begin{table}[ht]
\centering
\caption{Evaluation regarding the attention strategy. The proposed one outperforms the channel attention idea. The metrics correspond to the average of all devices.}
\begin{tabular}{@{}llll@{}}
\toprule
          & PSNR $\uparrow$  & \dE $\downarrow$  & SSIM $\uparrow$  \\ \midrule
Proposed attention (D) & \textbf{26.16} & \textbf{6.06} & \textbf{0.807}   \\ 
Channel attention & 25.44 & 6.35 & 0.791  \\  \midrule
\end{tabular}

\label{tab:attn}
\end{table}

\begin{table}[ht]
\centering
\caption{\model evaluates if the pre-training with monitor data and the frozen statistics strategy were removed. The metrics correspond to the average of all devices.}
\begin{tabular}{@{}llll@{}}
\toprule
          & PSNR $\uparrow$  & \dE $\downarrow$  & SSIM $\uparrow$  \\ \midrule
- Pre-training Monitor    & 24.96     &  7.07   & 0.782   \\ 
- Frozen Normalization & 26.22 & 6.26 & 0.811  \\
All Contributions (E) & \textbf{26.63}      &  \textbf{5.91} &  \textbf{0.817}       \\ \midrule
\end{tabular}

\label{tab:norm}
\end{table}

Table~\ref{tab:attn} and \ref{tab:norm} demonstrate that the proposed attention outperforms channel attention idea and the normalization technique impact together with pre-training with monitor data. Table~\ref{tab:individual} demonstrate that our method
performs well even when trained individually for each device. We observed that iPhone XR accuracy showed a slight
improvement. Meanwhile, the Pixel 6 Pro and Samsung S22 Ultra decreased
their metrics. This shows that an individual translation between the
raw image and its RGB version is more accurate when tackled
alone. However, when translating and processing images from different
devices, the jointly learned version has the advantage of having
access to various color possibilities, serving as a kind of data
augmentation that enriches the training. Finally, Table~\ref{tab:param} shows the parameter counting for all proposed models.

\begin{table}[ht]
\centering
\caption{\model also achieves state-of-the-art results when trained to independently translate from iPhone raw to Pixel, Samsung, and iPhone color appearances. }
\begin{tabular}{@{}llll@{}}
\toprule
          & PSNR $\uparrow$  & \dE $\downarrow$  & SSIM $\uparrow$  \\ \midrule
Iphone XR   & 31.64 & 3.13 & 0.931     \\ 
Pixel 6 Pro & 24.80 & 7.19 & 0.780        \\ 
Samsung S22 Ultra& 23.56 & 7.82 & 0.742 \\\midrule
\end{tabular}

\label{tab:individual}
\end{table}

\begin{table}[ht]
\centering
\caption{\model compared against SOTA in terms of number of parameters.}
\begin{tabular}{@{}lr@{}}
\toprule
          & Parameters (M) $\downarrow$ \\ \midrule
MW-ISPnet~\cite{contest2020} &     29.2           \\
LiteISP~\cite{zhangliteisp}   &     11.9           \\
SwinIR~\cite{liang2021swinir}    &     11.8           \\
\textbf{\model} &       \textbf{6.4}         \\ \midrule
\end{tabular}
\label{tab:param}
\end{table}

\section{Additional Results}
In this section, we present additional visualizations that include the results of illuminant estimation as well as failure cases and patch visualizations of the translation process from raw images captured with an iPhone XR to Pixel 6 Pro, Samsung S22 Ultra, and iPhone XR devices. Lastly, we showcase zero-shot examples of image translation from raw images taken with a Xiaomi C40.

\subsection{Illuminants Estimation}

In Figure~\ref{eq:illu}, the left side illustrates cases where the illuminant estimation outperforms the use of iPhone white balance information. This is because the white balance training is based on the original scalar values of the target phone, which can result in improved color accuracy for Pixel and Samsung devices. Without this strategy, the network relies solely on learning the illuminant transfer without any hint of the original values, which may lead to suboptimal results. This approach may fails in challenging lighting conditions that are not well represented in the training dataset, resulting in inaccuracies during inference.

\subsection{Additional Visualizations}

\begin{figure*}[h]
\centering
\includegraphics[width=0.7\linewidth]{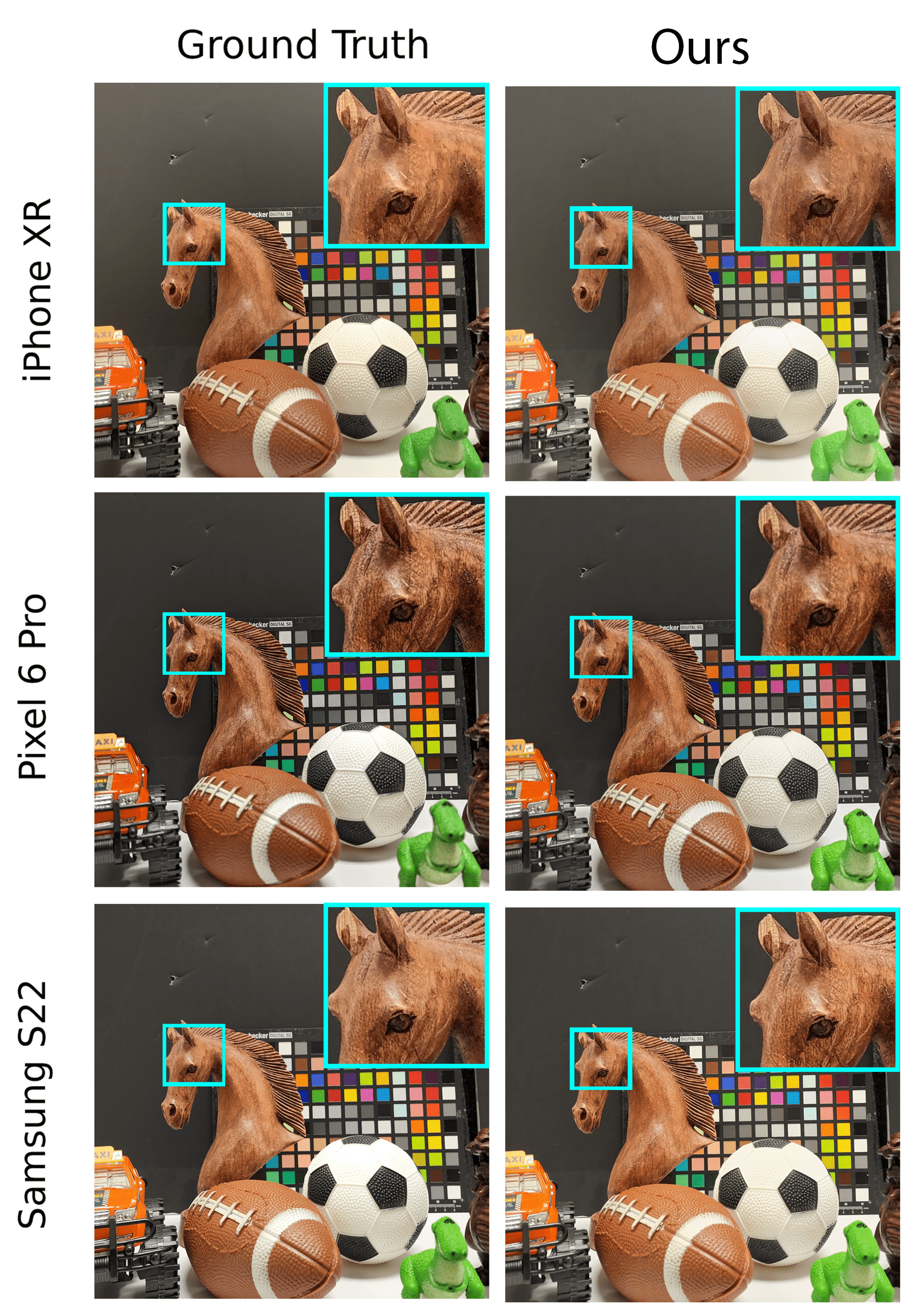}
\caption{Full-resolution inference visualization. Here, we highlight the details of the horse that are more apparent in the Pixel images. \model can accurately reproduce the color perception using the raw input from the iPhone, showcasing its ability to generate visually pleasing and accurate results, even in cases where the color rendition of different devices varies. This demonstrates the effectiveness of our approach in preserving and reproducing device-specific color characteristics, as evident in the high-quality output visualizations.}
\label{fig:resintro}
\end{figure*}

\begin{figure*}[h]
\centering
\includegraphics[width=1.0\linewidth]{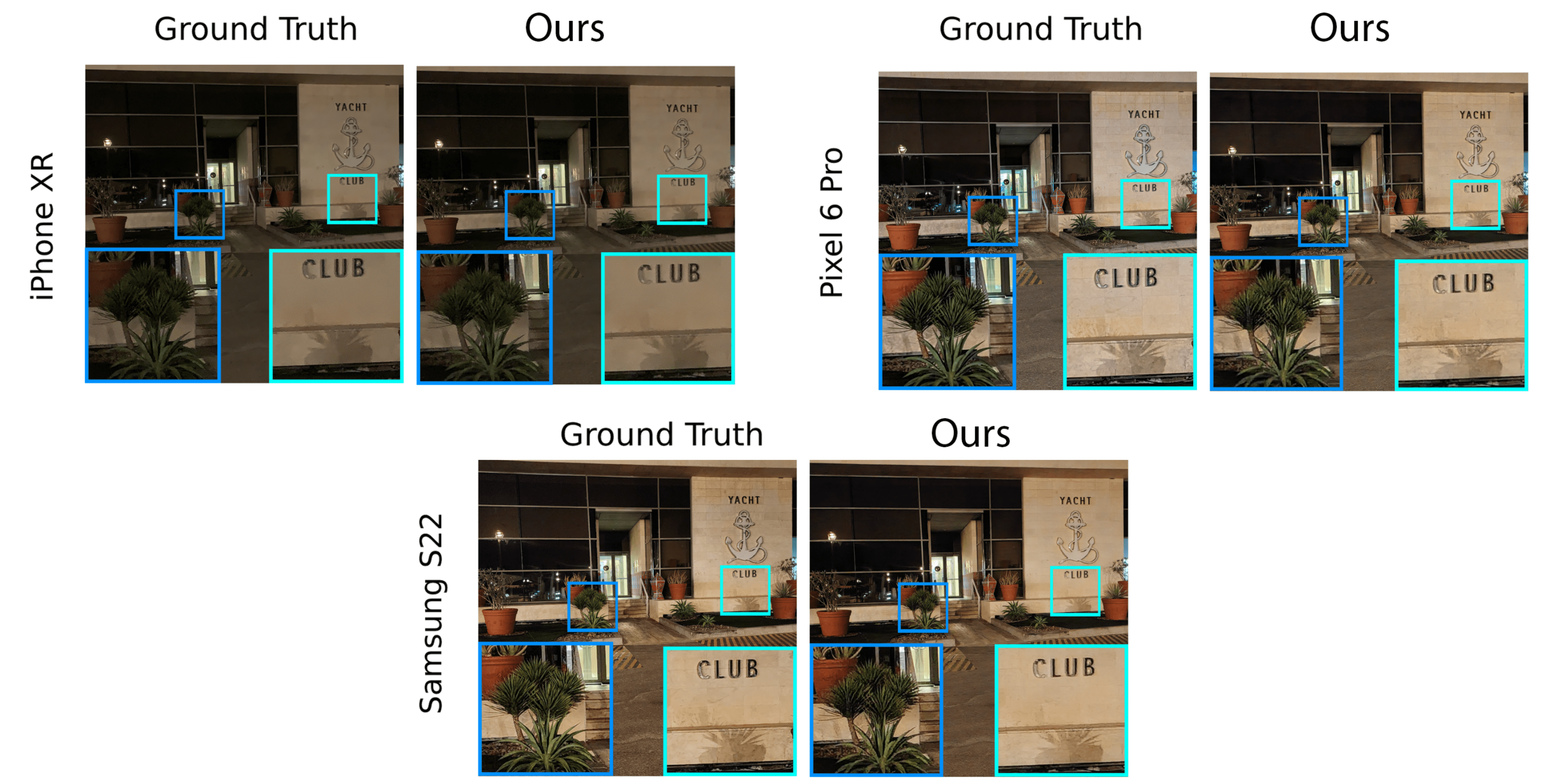}
\caption{Full resolution with comparison - Night Scene.}
\label{fig:p2}
\end{figure*}

\begin{figure*}[h]
\centering
\includegraphics[width=0.7\linewidth]{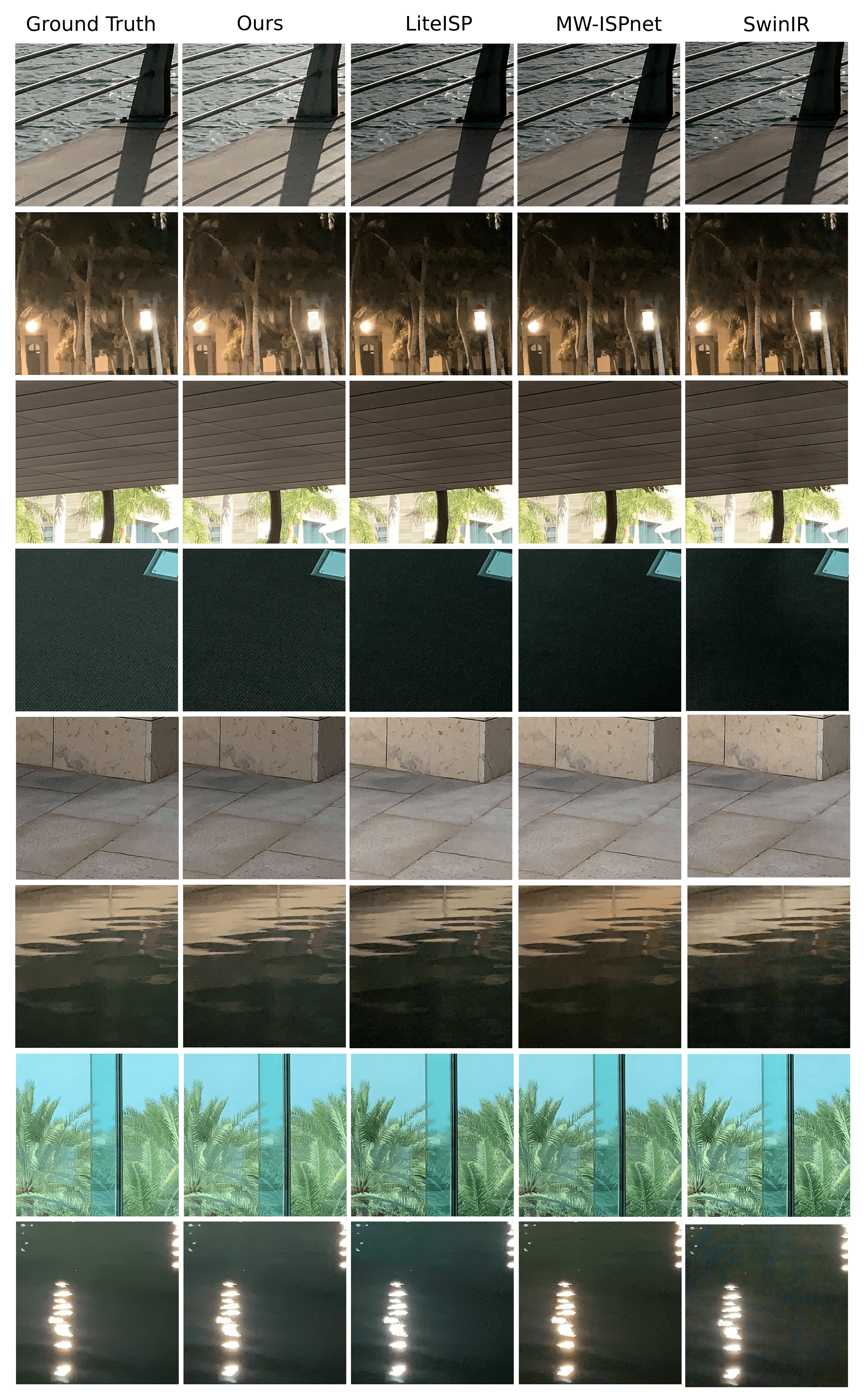}
\caption{Patch comparison - iPhone XR.}
\label{fig:iphone}
\end{figure*}

\begin{figure*}[h]
\centering
\includegraphics[width=0.7\linewidth]{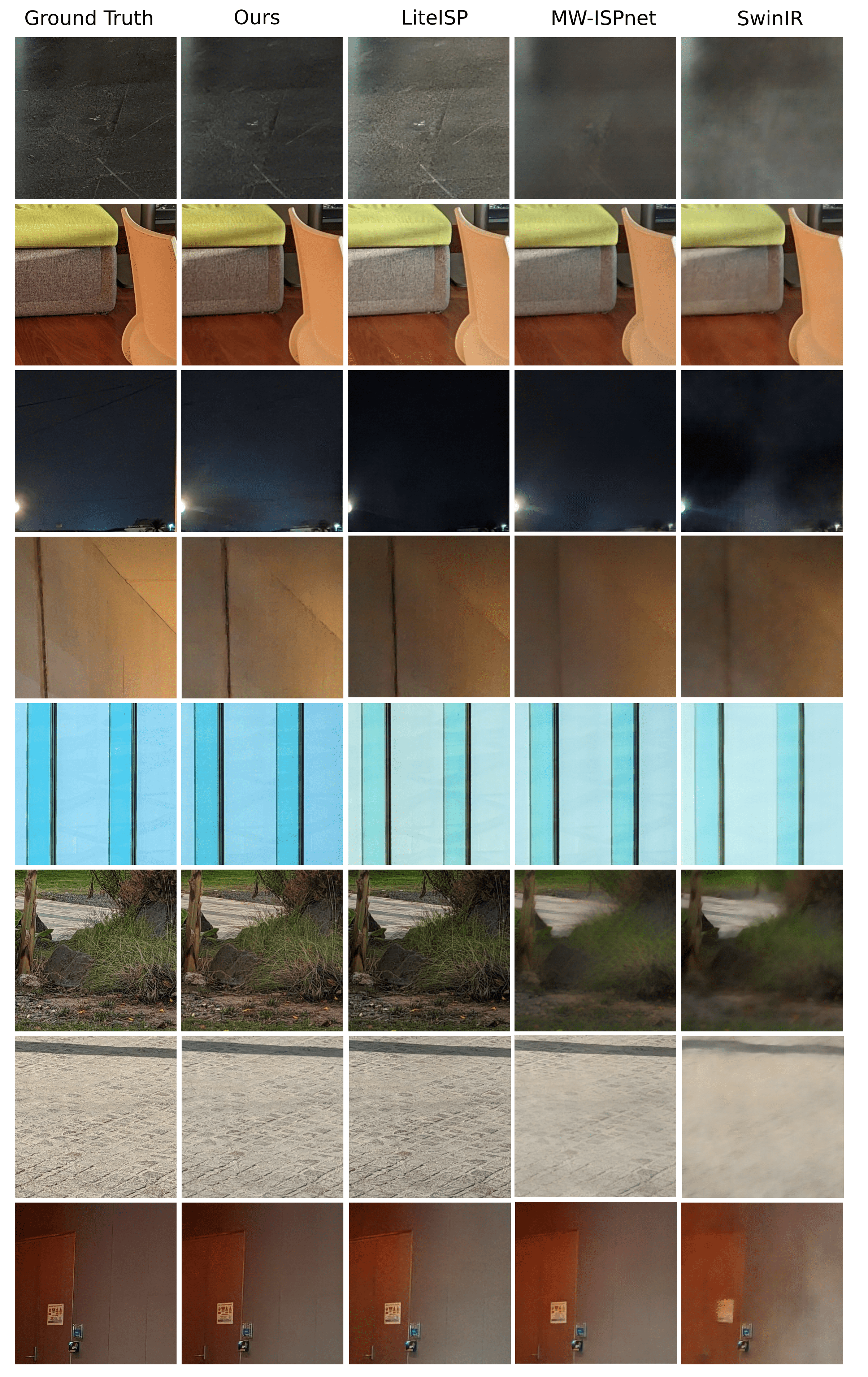}
\caption{Patch comparison - Pixel 6 Pro.}
\label{fig:pixel}
\end{figure*}

\begin{figure*}[h]
\centering
\includegraphics[width=0.7\linewidth]{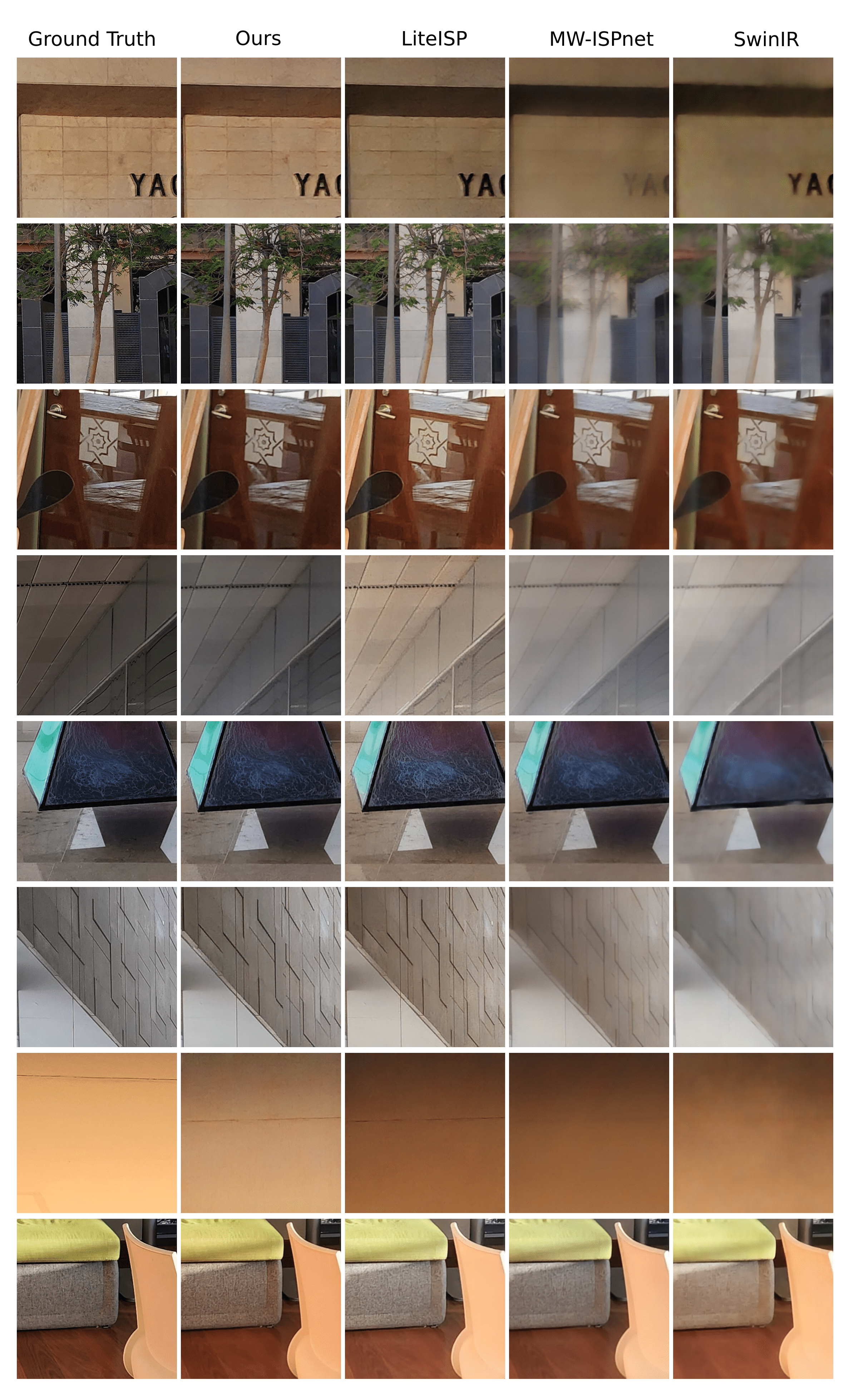}
\caption{Patch comparison - Samsung S22 Ultra.}
\label{fig:samsung}
\end{figure*}

Figures~\ref{fig:iphone},\ref{fig:pixel}, and \ref{fig:samsung} present detailed qualitative comparisons, showcasing the color accuracy of our model across various lighting conditions, such as night scenes, indoor settings, and outdoor environments, for all target devices. In Figure~\ref{fig:p2} and  \ref{fig:resintro}, we provide an additional full-resolution image visualization, comparing our results against the ground truth. These visualizations offer comprehensive insights into the performance of our model in different scenarios, highlighting its effectiveness in achieving accurate color reproduction.

\subsection{Zero-shot}

\begin{figure*}[h]
\centering
\includegraphics[width=1.0\linewidth]{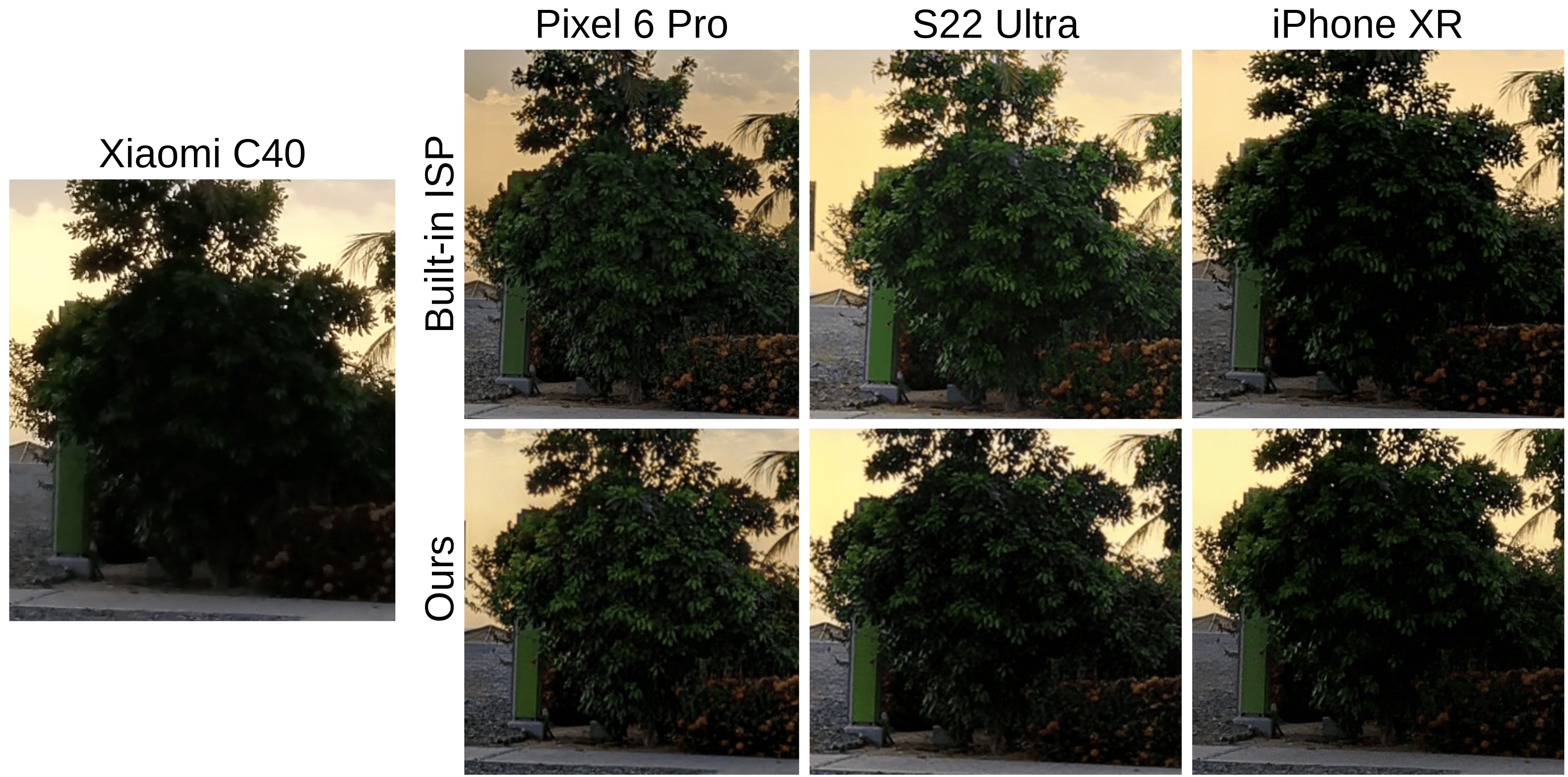}
\caption{Xiaomi C40 - 1}
\label{fig:tree}
\end{figure*}

\begin{figure*}[h]
\centering
\includegraphics[width=1.0\linewidth]{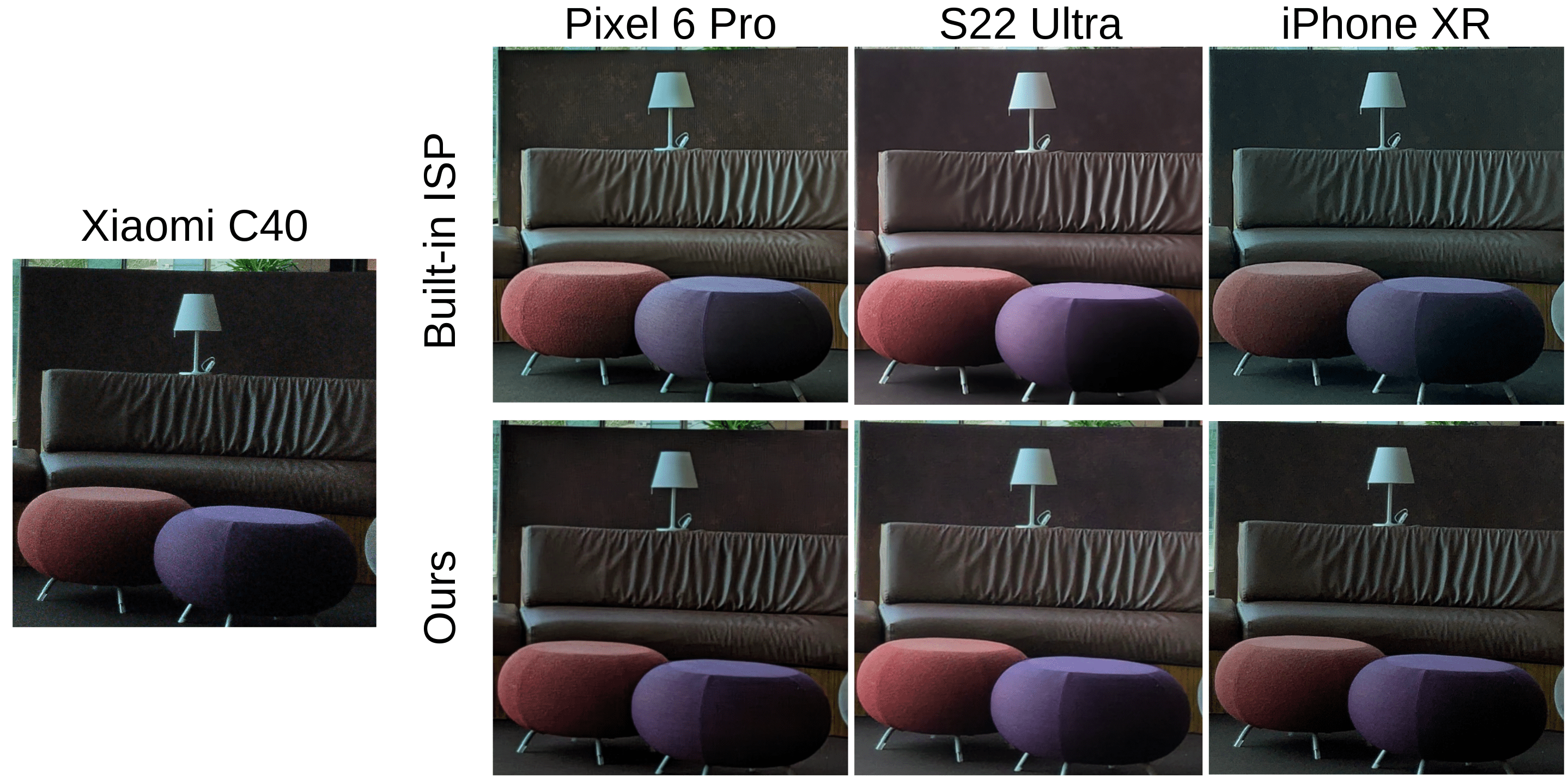}
\caption{Xiaomi C40 - 2}
\label{fig:seat}
\end{figure*}

\begin{figure*}[h]
\centering
\includegraphics[width=0.6\linewidth]{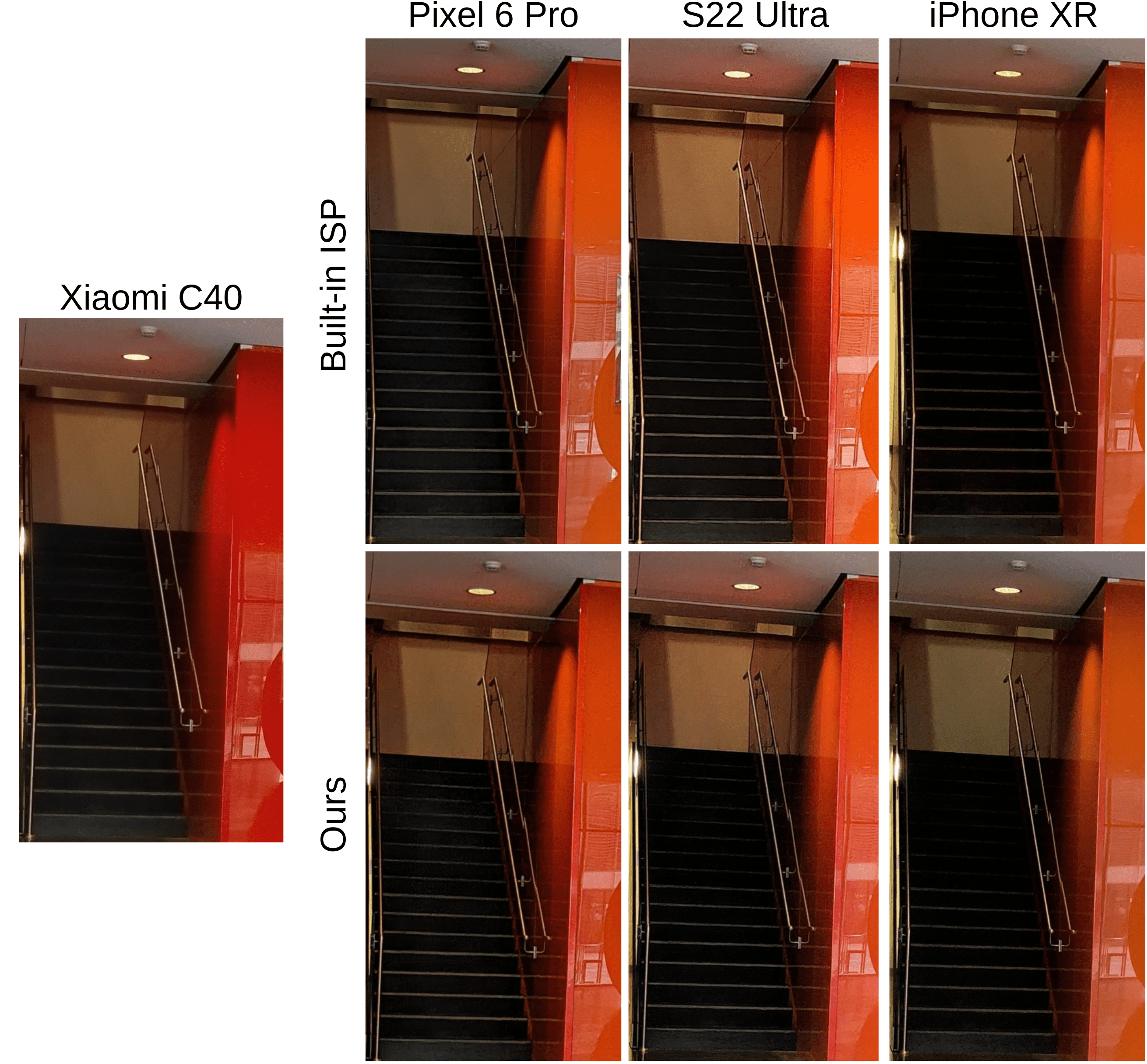}
\caption{Xiaomi C40 - 3}
\label{fig:red}
\end{figure*}

\begin{figure*}[h]
\centering
\includegraphics[width=0.7\linewidth]{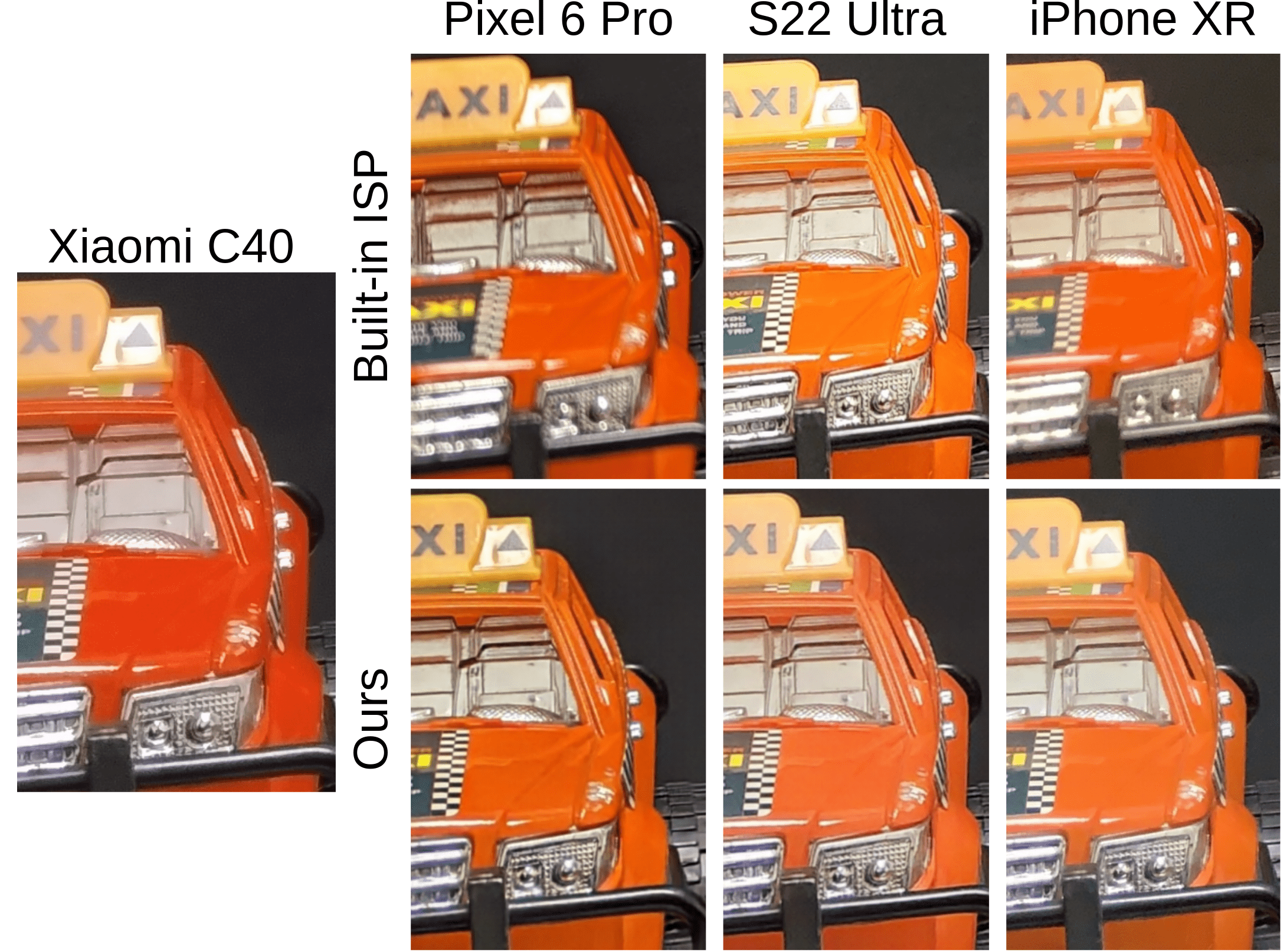}
\caption{Xiaomi C40 - 4}
\label{fig:car}
\end{figure*}

\begin{figure*}[h]
\centering
\includegraphics[width=1.0\linewidth]{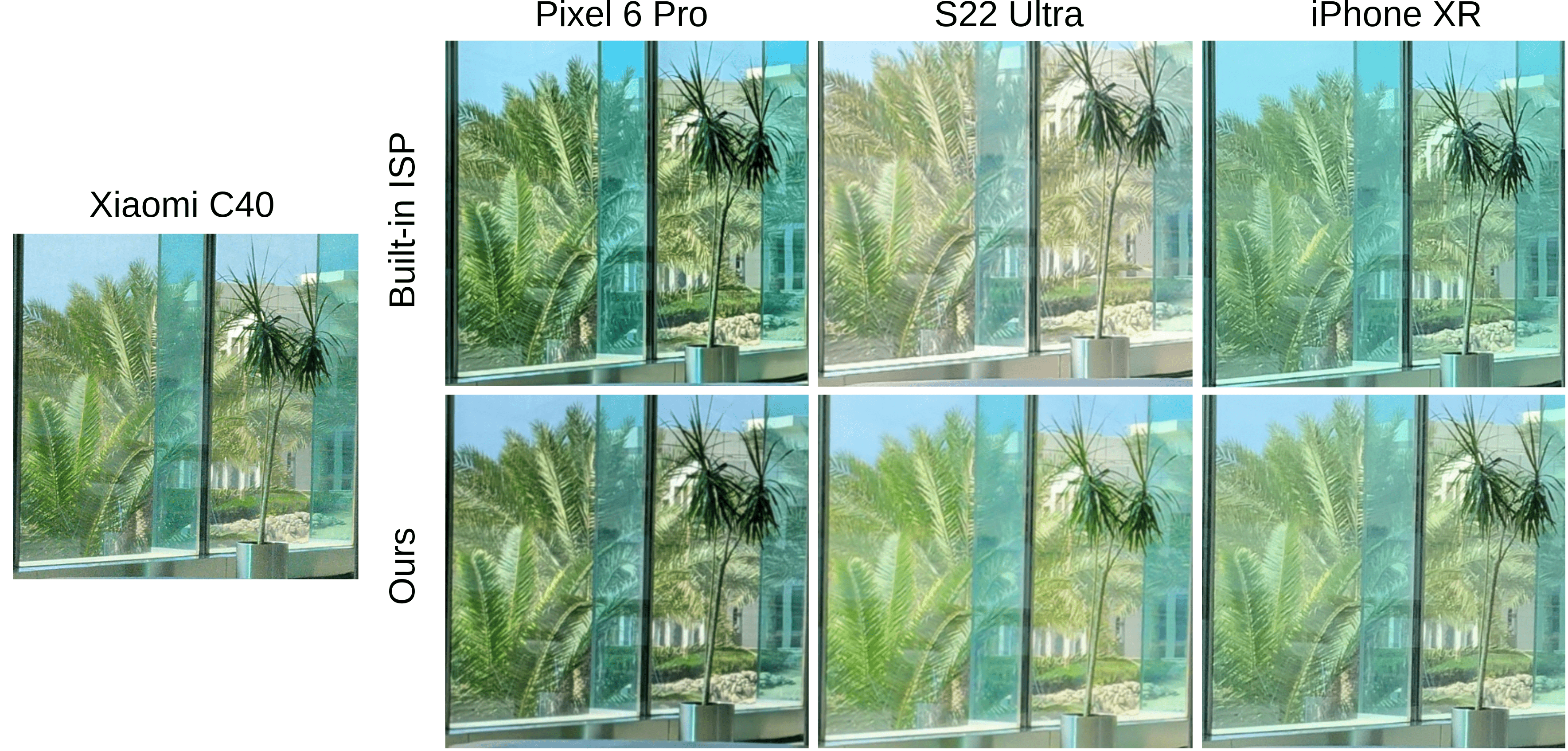}
\caption{Xiaomi C40 - 5}
\label{fig:lib}
\end{figure*}

\begin{figure*}[h]
\centering
\includegraphics[width=1.0\linewidth]{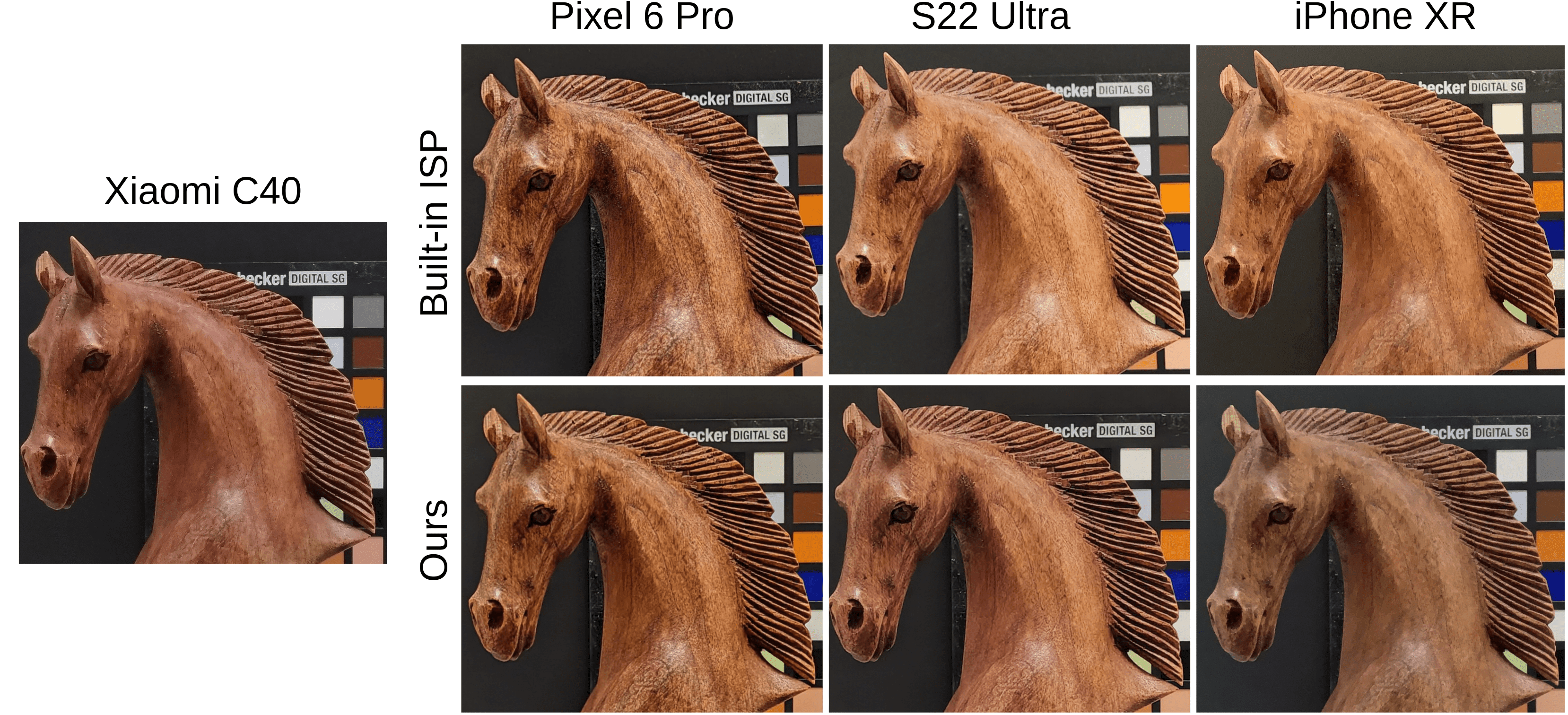}
\caption{Xiaomi C40 - 6}
\label{fig:horse}
\end{figure*}

Figures~\ref{fig:tree},\ref{fig:seat},\ref{fig:red},\ref{fig:car},\ref{fig:lib}, and~\ref{fig:horse} showcase additional examples of our method's generalization capabilities using raw images from Xiaomi C40 without further training. 

\begin{figure*}[h]
\centering
\includegraphics[width=1.0\linewidth]{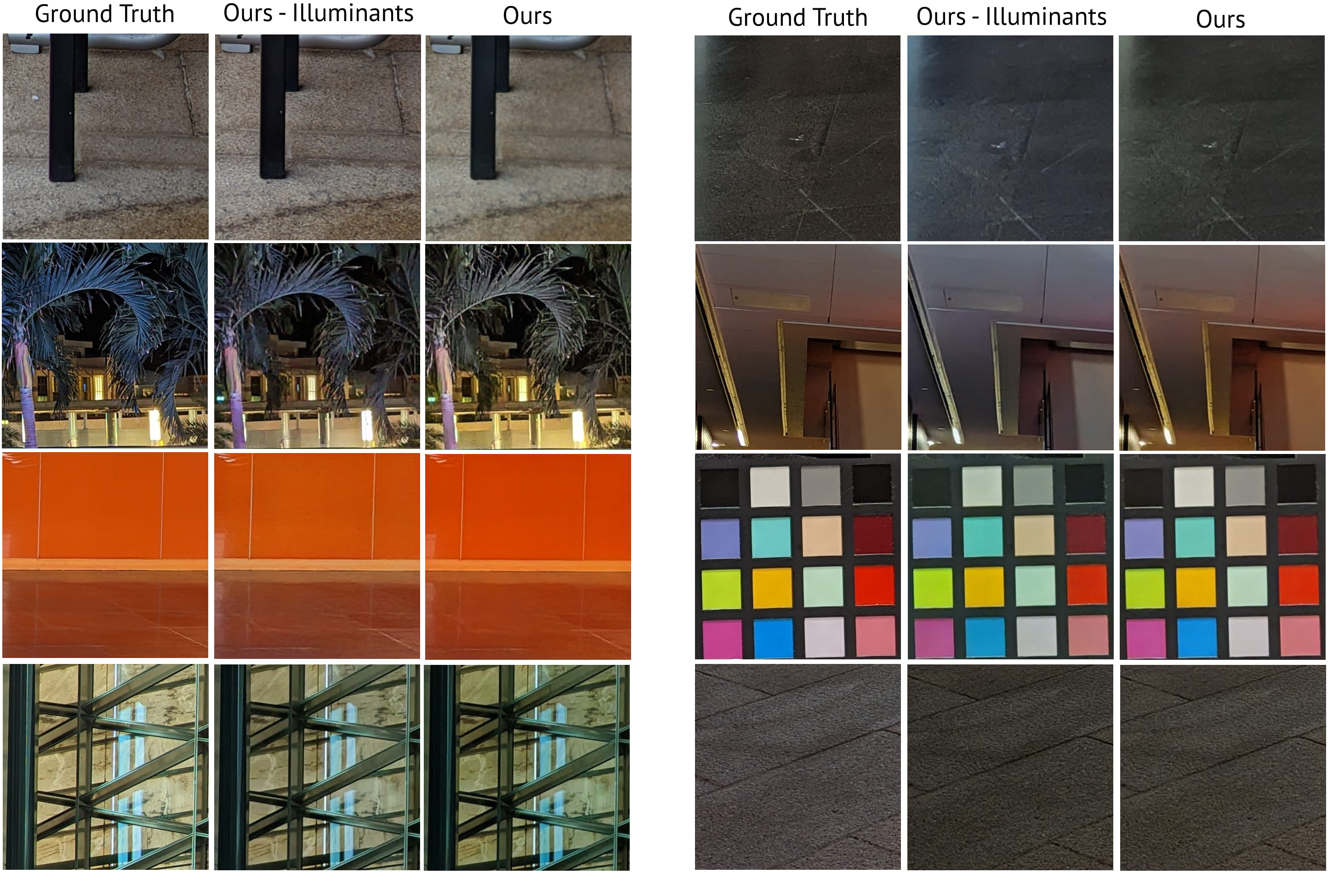}
\caption{The illuminants analysis is depicted in the images on the left side, which illustrate examples where the estimation of illuminants aids in achieving more accurate colors. On the right side, we present some failure cases where the estimation layer is unable to provide accurate white balance information, resulting in color deviations.}
\label{fig:illu}
\end{figure*}

%-------------------------------------------------------------------------

\end{document}